\documentclass[conference]{IEEEtran}
\IEEEoverridecommandlockouts
\usepackage{cite}
\usepackage{float}
\usepackage{algpseudocode}
\usepackage{csquotes}
\usepackage{algorithm}
\usepackage{enumitem}
\usepackage{amsmath,amssymb,amsfonts}
\usepackage{array}
\usepackage{graphicx}
\graphicspath{ {./images/} }
\usepackage{subcaption}   % 用于子图布局
\usepackage{textcomp}
\usepackage{xcolor}
\usepackage{multirow}
\def\BibTeX{{\rm B\kern-.05em{\sc i\kern-.025em b}\kern-.08em
    T\kern-.1667em\lower.7ex\hbox{E}\kern-.125emX}}
    
\begin{document}
%\title{MFF-FTNet: Enhanced Contrastive Learning with Multi-Scale Frequency and Time Domain Feature for Time Series Forecasting\\
\title{MFF-FTNet: Multi-scale Feature Fusion across Frequency and Temporal Domains for Time Series Forecasting
%\title{ An Adaptive Frequency and Time  Domain Feature Fusion Neural Network for Tiem Series Forecasting
}
\author{\IEEEauthorblockN{Yangyang Shi$^1$, Qianqian Ren$^1$\thanks{$Corresponding \ author: Qianqian \ Ren.$ }, Yong Liu$^1$, Jianguo Sun$^2$}
\IEEEauthorblockA{$^1$Department of Computer Science and Technology, Heilongjiang University,Harbin, China\\
$^2$Hangzhou Institute of Technology, Xidian University, Hangzhou, China}
% \IEEEauthorblockA{$^2$Hangzhou Institute of Technology, Xidian University, Hangzhou, China}\\
syy12248@s.hlju.edu.cn, \{renqianqian, liuyong123456\}@hlju.edu.cn, jgsun@xidian.edu.cn
}
\maketitle

\begin{abstract} 
Time series forecasting is crucial in many fields, yet current deep learning models struggle with noise, data sparsity, and capturing complex multi-scale patterns. This paper presents MFF-FTNet, a novel framework addressing these challenges by combining contrastive learning with multi-scale feature extraction across both frequency and time domains. MFF-FTNet introduces an adaptive noise augmentation strategy that adjusts scaling and shifting factors based on the statistical properties of the original time series data, enhancing model resilience to noise. The architecture is built around two complementary modules: a Frequency-Aware Contrastive Module (FACM) that refines spectral representations through frequency selection and contrastive learning, and a Complementary Time Domain Contrastive Module (CTCM) that captures both short- and long-term dependencies using multi-scale convolutions and feature fusion. A unified feature representation strategy enables robust contrastive learning across domains, creating an enriched framework for accurate forecasting. Extensive experiments on five real-world datasets demonstrate that MFF-FTNet significantly outperforms state-of-the-art models, achieving a 7.7\% MSE improvement on multivariate tasks. These findings underscore MFF-FTNet’s effectiveness in modeling complex temporal patterns and managing noise and sparsity, providing a comprehensive solution for both long- and short-term forecasting.

\end{abstract}


\section{Introduction}
Time series forecasting is vital across numerous domains, providing critical insights by predicting future values based on historical data. This technique is essential in fields such as finance \cite{zhao2023doubleadapt, cao2022ai, houssein2022efficient}, weather prediction \cite{bi2023accurate}, energy management \cite{zhuang2023data}, traffic forecasting \cite{jiang2023spatio}, and healthcare \cite{morid2023time}.
Time series data are characterized by a variety of complex dynamics-trends, seasonality, cyclic patterns, and random fluctuations \cite{dudek2023std}. moreover, this data is frequently noisy and incomplete, making it challenging to develop models that can accurately capture these intricate features while maintaining robustness and prediction accuracy.

Deep learning has greatly advanced time series forecasting by providing models that can capture intricate temporal patterns. Recurrent Neural Networks (RNNs)\cite{tokgoz2018rnn, wang2022ngcu} and their variants, Long Short-Term Memory (LSTM) networks\cite{siami2019performance, yin2023u}, were among the first deep learning approaches applied to time series forecasting. In parallel, Convolutional Neural Networks (CNNs) have shown strength in extracting local patterns in time series data\cite{livieris2020cnn, jin2020prediction}.
Temporal Convolutional Networks (TCNs) have also gained traction due to their ability to capture long-range dependencies through dilated convolutions, while preserving computational efficiency \cite{yue2022ts2vec, fraikin2023t, zheng2023simts}. More recently, Transformer models have been adapted for time series forecasting, leveraging the self-attention mechanism to model long-term dependencies flexibly \cite{zhou2021informer, nie2022time}. However, despite these advancements, these models often struggle to handle noise, data sparsity, and the need to capture multi-scale patterns effectively.

\begin{figure}
     \centering
     \includegraphics[width =\linewidth]{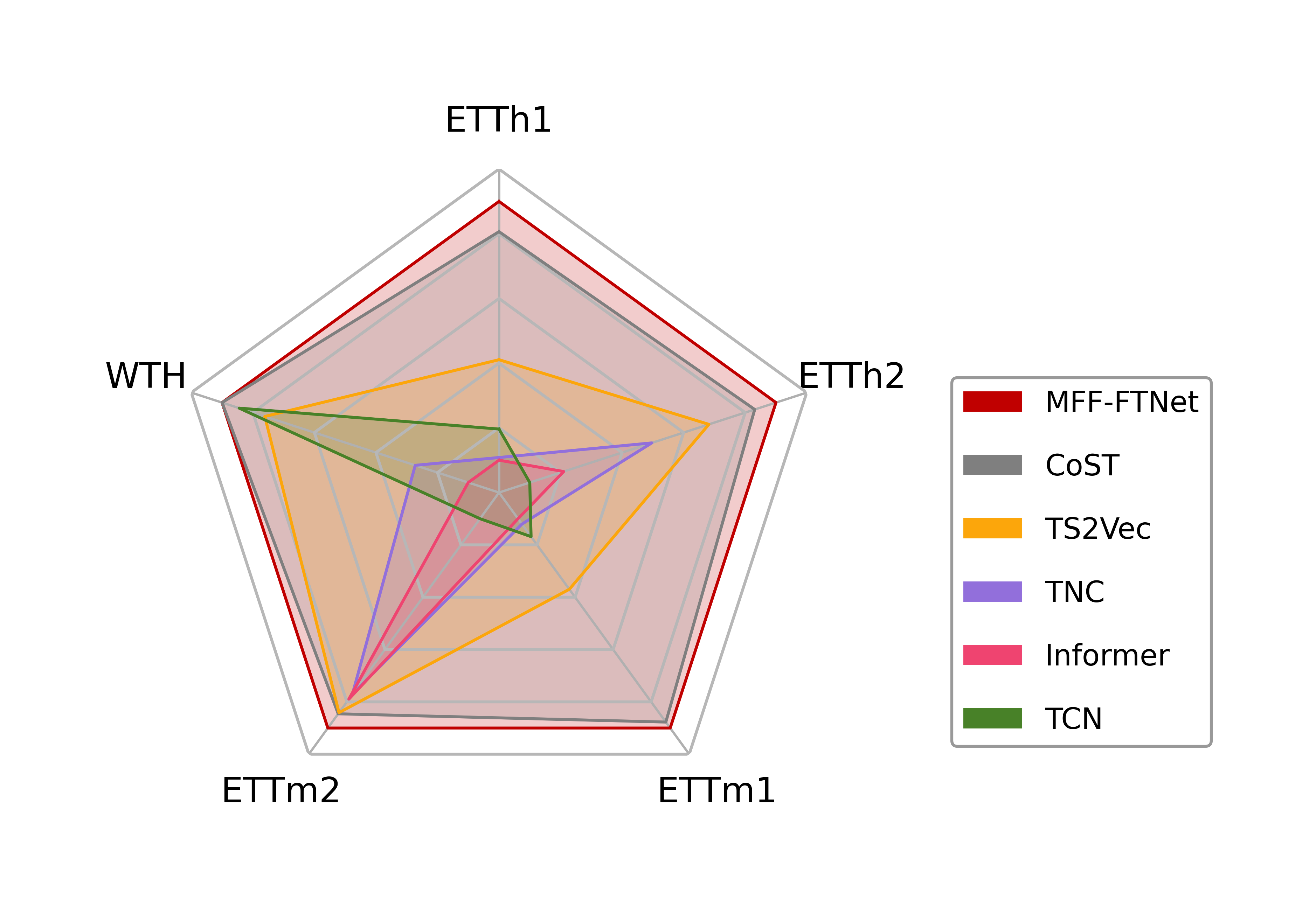}
     \caption{Performance comparisons of our model and other baseline models on five datasets. MFF-FTNet consistently achieved the best performance across all horizons.}
     \label{fig:radar}
 \end{figure}

To further enhance deep learning models for time series forecasting, contrastive learning has been introduced to improve representation quality by augmenting data and learning to differentiate meaningful patterns\cite{darban2025carla, woo2022cost, yue2022ts2vec}. By generating contrasting samples, contrastive learning methods can learn more robust features that are less sensitive to noise and data sparsity. For example, CoST \cite{woo2022cost} decomposes the time series into two independent components: seasonality and trend.  This decomposition allows CoST to design component-specific data augmentation strategies, generating contrasting samples tailored to each component and employing a dual contrastive loss to enhance representation learning. TS2Vec \cite{yue2022ts2vec} introduces a cross-temporal context contrastive learning objective, using data augmentation techniques such as temporal masking, cropping, and jittering to create augmented views of the time series. These contrasting views enable the model to learn a robust distinction between similar and dissimilar temporal features, thereby improving its understanding of complex time series patterns. 

Despite promising results in time series forecasting, there are three crucial aspects that have often been neglected.

\begin{itemize}
 \item \textbf{Limited Noise Resilience in Forecasting Models:}
Many state-of-the-art time series forecasting methods fail to effectively handle noisy or non-stationary data, which is common in real-world applications. Existing solutions may over-smooth the data or become overly sensitive to noise, impairing their forecasting accuracy. The proposed Frequency-Aware Contrastive Module (FACM) addresses this issue by enhancing the model’s robustness to noise through adaptive frequency selection and contrastive learning in the frequency domain.
\item \textbf{Inadequate Multi-Scale Pattern Capture:}
Existing time series forecasting models often struggle to effectively capture multi-scale patterns, especially those that span both short-term and long-term dependencies. Current methods may focus on either short-term fluctuations or long-term trends but fail to integrate both. This paper addresses this limitation by proposing a method that captures multi-scale temporal features, enabling the model to better understand and predict across different time horizons.
 \item \textbf{Inability to Integrate Time and Frequency Domain Features Effectively:}
While some models focus on either time-domain or frequency-domain features, few frameworks integrate both in a complementary manner. This results in a limited understanding of time series data, as they are unable to leverage the complementary strengths of both domains. This paper addresses this gap by combining time-domain and frequency-domain features, providing a more holistic representation of the data for improved forecasting.
\end{itemize} 

To address the challenges of capturing multi-scale patterns, handling noisy data, and integrating both time and frequency domain features, we propose a novel time series forecasting model, called \textit{Multi-scale Feature Fusion across Frequency Temporal Domains for Time Series Forecasting} (MFF-FTNet).
To address the issue of data noise (\textbf{issue 1}), we introduce the Frequency-Aware Contrastive Module (FACM), which enhances frequency domain representations by applying a masking operation to filter out irrelevant high-frequency noise. This enables the model to generate more robust and noise-resistant frequency-domain features, thereby improving forecasting accuracy. Furthermore, contrastive learning is applied to both the amplitude and phase components within the frequency domain, strengthening the model’s discriminative capabilities and adaptability to noisy data.
For the challenge of capturing multi-scale temporal dependencies (\textbf{issue 2}), we incorporate the Complementary Time Domain Contrastive Module (CTCM). This module employs multiple convolutions with varying kernel sizes to capture multi-scale temporal features and complex nonlinear patterns. By utilizing contrastive learning in both the time and frequency domains, the model identifies subtle yet essential differences across various time points and feature dimensions, thereby improving its ability to learn intricate temporal structures.
To solve the problem of integrating time and frequency domain features (\textbf{issue 3}), the model combines features from both domains through a unified feature representation strategy. This enables robust contrastive learning across time and frequency domains, providing a comprehensive and enriched forecasting framework that enhances model performance.

In summary, the main contributions of this work are summarized as follows:
\begin{itemize}
\item \textbf{General Aspect.}  
This paper presents MFF-FTNet, a novel framework that combines contrastive learning with multi-scale frequency and time-domain feature extraction for time series forecasting. By integrating frequency-aware and time-domain complementary modules, MFF-FTNet effectively captures both high-level periodic patterns and localized temporal details, addressing key challenges such as noise resilience, multi-scale pattern detection, and robust forecasting under complex dependencies.

\item \textbf{Methodologies.}  
MFF-FTNet introduces two complementary modules to enhance feature extraction and representation. The Frequency-Aware Contrastive Module (FACM) refines spectral representations using adaptive frequency selection and dual-frequency contrastive loss, enhancing noise tolerance and adaptability to non-stationary data. The Complementary Time Domain Contrastive Module (CTCM) leverages multi-scale convolutions and feature fusion to capture intricate temporal dependencies across varying scales. Additionally, the model employs a combined feature representation strategy for robust contrastive learning across both domains, providing a unified and enriched forecasting framework.

\item \textbf{Experimental Evaluation.}  
 Extensive experiments on five real-world datasets validate the effectiveness of MFF-FTNet. As illustrated in Figure \ref{fig:radar}, MFF-FTNet consistently surpasses state-of-the-art baselines, achieving a 7.7\% reduction in MSE and a 4.0\% reduction in MAE for multivariate forecasting, and a 2.5\% reduction in MSE and 2.4\% in MAE for univariate forecasting.
\end{itemize}

\section{Related Work}
\subsection{Time Series Forecasting}
Time series forecasting has garnered significant attention in recent years due to its applications across various domains. Methods based on Graph Neural Networks (GNNs) have shown promising results in capturing complex dependencies in multivariate time series data. For instance, FourierGNN \cite{yi2024fouriergnn} introduces a Fourier-based graph neural network architecture that performs matrix multiplication in Fourier space, achieving both high expressive power and reduced complexity for effective and efficient predictions. CrossGNN \cite{huang2023crossgnn} addresses noise in time series by using adaptive multi-scale identifiers, constructing denoised multi-scale time series, and leveraging the homogeneity and heterogeneity between variables. This approach reduces time and space complexity linearly with the input series length. Although these models improve computational efficiency, they may still struggle to capture long-term dependencies across different time scales, which are crucial for accurate forecasting in many applications.

Transformer-based models have been successful in time series prediction especially for long term prediction. Crossformer \cite{zhang2023crossformer} incorporates cross-time and cross-dimensional models to capture correlations between multivariate time series. PatchTST \cite{nie2022time} divides time series into subseries-level patches and trains the model in a channel-independent manner, improving performance in time series tasks.
Despite the success of transformer-based approaches, the need for high computational resources and large amounts of labeled data also limits their scalability and applicability to real-world scenarios. Furthermore, many transformer models do not fully exploit the complementary information between the time and frequency domains, which can improve model robustness and prediction accuracy. 

\subsection{Contrastive Learning for Time Series Forecasting.}
Contrastive learning has gained significant attention for time series representation learning \cite{chen2020simple, xie2020unsupervised, you2020graph}, which constructs positive and negative sample pairs to learn data representations by minimizing the distance between similar samples and maximizing the distance between dissimilar ones. Contrastive learning-based deep neural networks have shown notable success in computer vision and natural language processing and time series representation and forecasting.
SimCLR \cite{chen2020simple} applies data augmentations (e.g., cropping, distortion, jitter) to generate enhanced views of the same image, using the NT-Xent loss function to optimize the network. MoCo \cite{he2020momentum} introduces a momentum update encoder and dynamic dictionary for negative samples, improving small batch training by maintaining diversity and consistency.
In the context of time series, TS2Vec \cite{yue2022ts2vec} employs contrastive learning on multiple time scales to capture multi-level features of time series. TS-TCC \cite{eldele2021time} generates positive sample pairs from sub-series and negative pairs from unrelated time series, learning both local and global features by maximizing the similarity of positive pairs. TimesURL \cite{liu2024timesurl} leverages self-supervised contrastive learning with time masking, jitter, and clipping augmentations, focusing on hard negative samples to improve model robustness.
However, despite the potential of contrastive learning in time series representation, several challenges remain. First, the definition of positive and negative pairs is non-trivial due to the temporal dependencies inherent in time series data, which can lead to less effective learning if not carefully constructed. Moreover, the reliance on data augmentation techniques such as time masking and jitter may not always generate realistic or meaningful variations of the time series, which can undermine the learning process. 

\section{METHODOLOGY}
\subsection{Problem Definition}
\textbf{Time Series Forecasting:} Consider a multivariate time series set $\mathbf{X} =\{\mathbf{x}_{1}, \mathbf{x}_{2}, \ldots, \mathbf{x}_{T}\} \in \mathbb{R}^{T \times D}$, where 
$T$ denotes the length of the observed window, with each time point characterized by 
$D$ distinct feature values. 
The objective of time series forecasting is to learn a mapping function $g(\cdot)$ such that the forecasting output $\mathbf{X}_O = g(\mathbf{X}) \in \mathbb{R}^{P \times D}$ accurately estimates the subsequent values of the series over the prediction horizon $P$.
%The goal of time series forecasting is to predict a future output series $\mathbf{X}_{O} = [\mathbf{x}_{L+1}, \mathbf{x}_{L+2}, \ldots, \mathbf{x}_{L+P}] \in \mathbb{R}^{P \times D}$, with $P$ representing the length of the prediction series. Supposed that the instance $\mathbf{x}_{i} \in \mathbb{R}^{T \times D} = \{x^{i,1}, \dots, x^{i,t}, \dots, x^{i,T}\}$ contains feature component vectors $\mathbf{x}^{i,t} \in \mathbb{R}^{D}$, where  $i\in \{1,2,\cdots,N\}$, $N$ is the njkkce. $\mathbf{x}_{i} $ is used as input and fed into the backbone module to obtain the output $\mathbf{r}^{i}\in \mathbb{R}^{T \times K}$. Simultaneously, $\mathbf{\tilde{h}}^{i} \in \mathbb{R}^{T \times \frac{K}{2} }$ and $\mathbf{\hat{h}}^{i} \in \mathbb{R}^{T \times \frac{K}{2} }$ are respectively represented as the time-domain and frequency-domain feature representations corresponding to $\mathbf{x}_{i}$. 
Additional symbol definitions can be found in Table 1.
\setlength{\extrarowheight}{2pt}
\begin{table}
    \caption{Description of notations.}
    \centering
    \begin{tabular}{l p{6.5cm}}
        \hline
        Notation & Description   \\ \hline
       % N & Number of input instances. \\
        T & Length of the input time series. \\
        D & Dimension of the time series data. \\
        P & Length of the prediction series. \\
        K & Dimension of the latent space vector. \\ \hline
        % $\mathbf{x}^{i}$ & Time series of instance $i$. \\
        $\mathbf{r}$ & Output of the backbone module using $\mathbf{X}$ as the input. \\
        $\mathbf{\tilde{h}}$ & Output of the CTCM using $\mathbf{r}$ as the input. \\
        $\mathbf{\hat{h}}$ & Output of the FACM using $\mathbf{r}$ as the input. \\
        $\mathbf{h}$ & Fusion of time-domain and frequency-domain feature matrices. \\ \hline

        $\mathbf{x}_t$ & A slice of $\mathbf{X}$ at a specific time step $t$, containing information on all features at that time step. \\
        $r_t$ & The representation of $\mathbf{x}_t$. \\
        $\tilde{h}_t$ & The time domain vector representation of $r_t$.\\
        $\hat{h}_t$ & The frequency domain vector representation of $r_t$.\\ 
        $h_t$ & The concatenation of $\tilde{h}_t$ and $\hat{h}_t$.\\ \hline 
        $L^ \prime$ & Number of dilated convolution blocks in Backbone.\\
        $D^ \prime$ & Intermediate dimension of the Backbone.\\
        $D^ {\prime\prime}$ & Intermediate dimension of the projection block in CTCM.\\
        $W$ & Weight decay coefficient of the optimizer.\\
        $n$ & Number of convolutional kernels.\\
        $\omega_p$ & Weight matrix of frequency components. \\
        $\beta_p$ & Bias matrix of frequency components.\\ 
        ${\alpha}_m $ & Frequency component masking ratio. \\ \hline
    \end{tabular}
    \label{table:Notation}
\end{table}

 \begin{figure*}
     \centering
     \includegraphics[width = 0.85\textwidth]{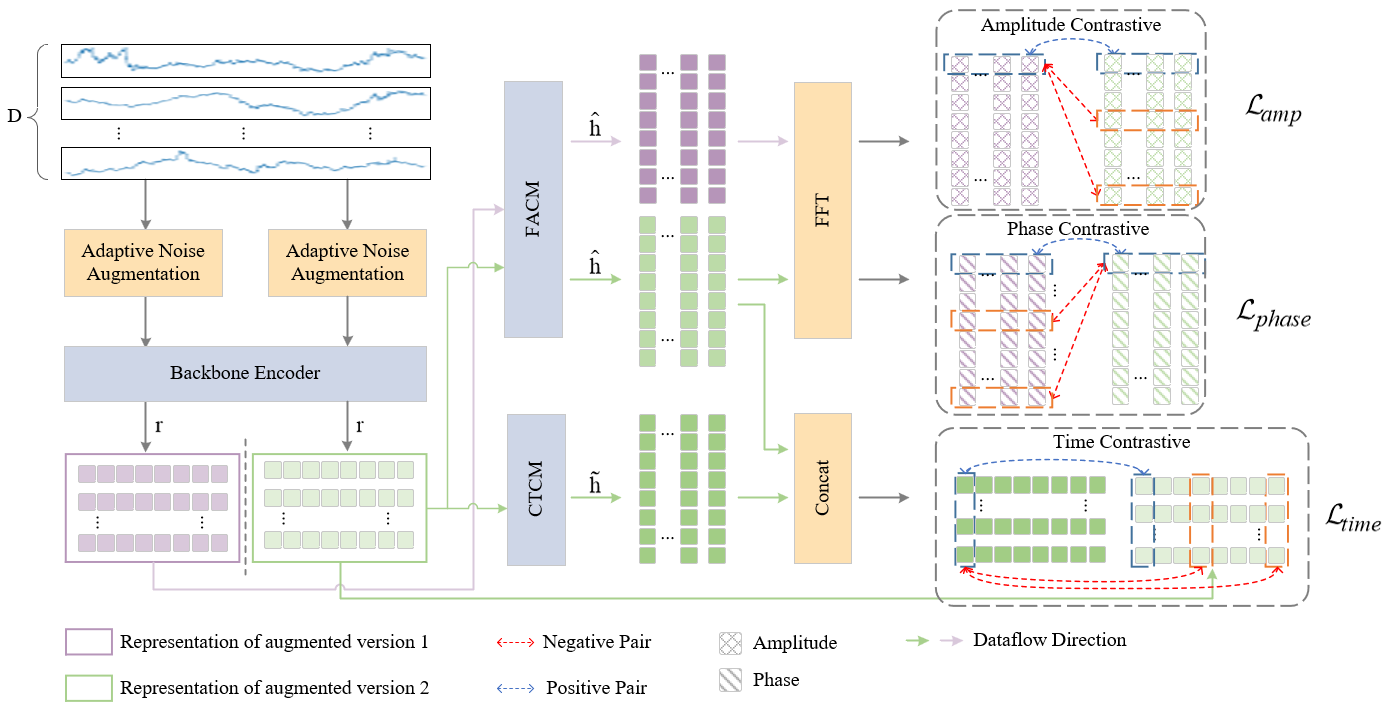}
     \caption{The proposed architecture of MFF-FTNet.}
     \label{figure:Overall FrameWork}
 \end{figure*}

\subsection{Model Architecture}
We provide an overview of the MFF-FTNet structure and functionality, with the complete architecture depicted in Figure \ref{figure:Overall FrameWork}.
The model starts by applying data augmentation techniques to the input time series data, generating two augmented views, which are then fed into the encoder module to latent space vectors for each view.
Next, the latent space vectors are separately passed to the Frequency-Aware Contrastive Module (FACM). In the FACM, the frequency domain features of the time series are extracted using Fast Fourier Transform (FFT), creating a complex frequency-domain matrix. A masking operation is then applied to remove noise, resulting in cleaner and more robust frequency-domain features. These features are used to compute the frequency domain loss.
Simultaneously, the latent space vector is sent to the Complementary Time Domain Contrastive Module (CTCM). The CTCM consists of multiple convolution layers with varying kernel sizes to capture multi-scale temporal features from the time domain. The corresponding frequency-domain features are concatenated with the time domain features, forming a richer feature representation. This combined representation is then used to calculate the feature loss, based on its similarity to the latent space vector.

\subsection{Adaptive Noise Augmentation}
%\textbf{Scaling:} It introduces variations in amplitude, which can help the model become invariant to changes in signal strength. Given a time series \( \mathbf{x} = \{x_1, x_2, \dots, x_T\} \), we apply a scaling factor to each time step:
%\begin{equation}
%\tilde{x}_t = \epsilon x_t
%\end{equation}
%where \( \epsilon \sim \mathcal{N}(1, \sigma^2) \). Here, we set \(\sigma\) to 0.5 to allow moderate variations around the original amplitude. Scaling by a Gaussian-distributed factor helps the model generalize to different signal intensities and makes it less sensitive to amplitude fluctuations.

%\textbf{Shifting:} It introduces a baseline offset, which can simulate environmental changes that alter the baseline of the signal. This augmentation is achieved by adding a random offset \( \epsilon \) to each time step:
%\begin{equation}
%\tilde{x}_t = x_t + \epsilon
%\end{equation}
%where \( \epsilon \sim \mathcal{N}(0, \sigma^2) \), with \( \sigma \) set to 0.5 in this case. This transformation makes the model less sensitive to baseline shifts, enabling it to focus more on relative changes rather than absolute values, which is particularly beneficial in applications like anomaly detection and trend analysis.

%With both scaling and shifting applied independently with a probability of 0.5, each transformed time series can be described as:
%\begin{equation}
%\tilde{x}_t = \epsilon_s \cdot x_t + \epsilon_b
%\end{equation}
%where \( \epsilon_s \sim \mathcal{N}(1, 0.5) \) and \( \epsilon_b \sim \mathcal{N}(0, 0.5) \) represent the scaling and shifting factors, respectively.

Adaptive noise augmentation dynamically adjusts the scaling and shifting factors based on the statistical properties of the original time series data. By tailoring these factors to the data's own mean and standard deviation, this technique generates augmentations that reflect the natural variations in amplitude and baseline shifts, enhancing the model’s ability to generalize.

Given a time series \( \mathbf{X} = \{\mathbf x_1, \mathbf x_2, \dots, \mathbf x_T\} \), we compute the mean $\mu_x$ and standard deviation $\sigma_x$ as follows:
 \begin{equation}
\begin{split}
   \mu_x &= \frac{1}{T} \sum_{t=1}^T \mathbf x_t\\
   \sigma_x &= \sqrt{\frac{1}{T} \sum_{t=1}^T (\mathbf x_t - \mu_x)^2}
   \end{split}
\end{equation}

Using $\sigma_x$ as a measure of the data’s natural variability, we apply scaling and shifting transformations to generate an augmented time series $\tilde{\mathbf x}_t$:

\begin{equation}
\tilde{\mathbf x}_t = \epsilon_s \cdot \mathbf x_t + \epsilon_b
\end{equation}
where \( \epsilon_s \sim \mathcal{N}(1, \alpha \cdot \sigma_x) \) is a scaling factor drawn from a normal distribution centered at 1 with standard deviation proportional to \( \sigma_x \). \( \epsilon_b \sim \mathcal{N}(0, \beta \cdot \sigma_x) \) is an offset factor drawn from a normal distribution centered at 0 with standard deviation proportional to \( \sigma_x \). \( \alpha \) and \( \beta \) are hyperparameters controlling the strength of the scaling and shifting noise, respectively. 

\subsection{Backbone Encoder}
\begin{figure}
     \centering
     \includegraphics[width =\linewidth]{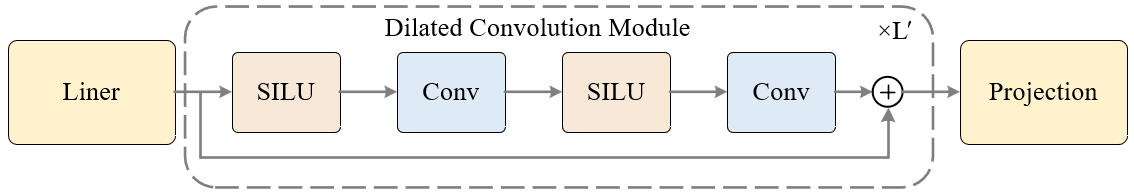}
     \caption{The architecture of the backbone encoder.}
     \label{fig: Backbone Encoder}
 \end{figure}

The augmented data is input to an encoder module based on the CoST backbone \cite{woo2022cost}, which encodes the time series data into latent feature vectors. As shown in Figure \ref{fig: Backbone Encoder}, this encoder consists of three main components: a linear layer, multiple dilated convolution blocks, and a projection layer.

First, we use a linear layer to map the input time series data $\tilde{\mathbf X}\in \mathbb{R}^{T \times D}$ into a high-dimensional latent space, capturing more feature information as a foundation for subsequent processing. Then, multiple dilated convolution blocks with residual connections are employed to extract features from the high-dimensional data. These dilated convolution modules effectively expand the receptive field, enabling the capture of long-term dependencies without increasing computational complexity, while the residual connections help maintain gradient stability and prevent information loss. Finally, the extracted features are mapped to the final output $\mathbf{r} \in \mathbb{R}^{T \times K}$ through the projection layer. The backbone encoder can be expressed as follows:
\begin{equation}
\mathbf{r}=\text{Projection(DilatedConv(Linear}(\tilde{\mathbf X})))
\end{equation}

It is noteworthy that we have replaced the GELU activation function with the SiLU (Sigmoid Linear Unit) activation function due to the advantages that SiLU provides in our context. SiLU has been demonstrated to yield smoother gradients and enhanced performance across various deep learning tasks, facilitating improved convergence during training and increasing model robustness.
Moreover, the non-monotonic nature of SiLU allows for greater flexibility in learning complex data patterns, rendering it particularly well-suited for tasks involving time series analysis. In our experiments, we observed that this modification resulted in a superior representation of temporal features, ultimately contributing to more accurate and reliable latent feature extraction.

\subsection{Frequency-Aware Contrastive Module(FACM)}
\subsubsection{Motivation}
Frequency domain feature extraction in time series forecasting has garnered significant attention in recent years due to its ability to reveal important periodic and spectral characteristics within complex temporal data\cite{woo2022cost,WuHLZ0L23,dai2024periodicity}. Traditional time domain analyses often struggle with intricate periodicities and overlapping trends, particularly in data with noise, seasonality, or multi-scale patterns. By transforming time series data into the frequency domain, critical cyclic components become more prominent.
%, enabling the model to identify and leverage recurring patterns that might otherwise be obscured. 
Additionally, frequency domain features can facilitate noise reduction. %enhance model interpretability, and allow for a more compact representation of data, leading to improved forecasting accuracy and computational efficiency.
%However, existing frequency domain feature extraction methods often based on the assumption of stationary or stable periodic characteristics, limiting their effectiveness in real-world applications where temporal patterns and frequencies shift over time. 
%This limitation results in poor adaptability to non-stationary data and reduces forecasting accuracy in dynamic environments. 
% Additionally, the transformation of data to the frequency domain may lead to a loss of critical temporal information, as frequency components do not inherently capture sequence order or recent event impacts. 
%Furthermore, current methods may struggle with selecting relevant frequency bands amidst noise or irrelevant components, which can hinder the model’s robustness. Finally, most existing approaches do not effectively integrate frequency domain features with modern deep learning frameworks, particularly contrastive learning, which could offer substantial benefits in refining representations and capturing meaningful patterns in complex data.

\begin{figure}
     \centering
     \includegraphics[width =\linewidth]{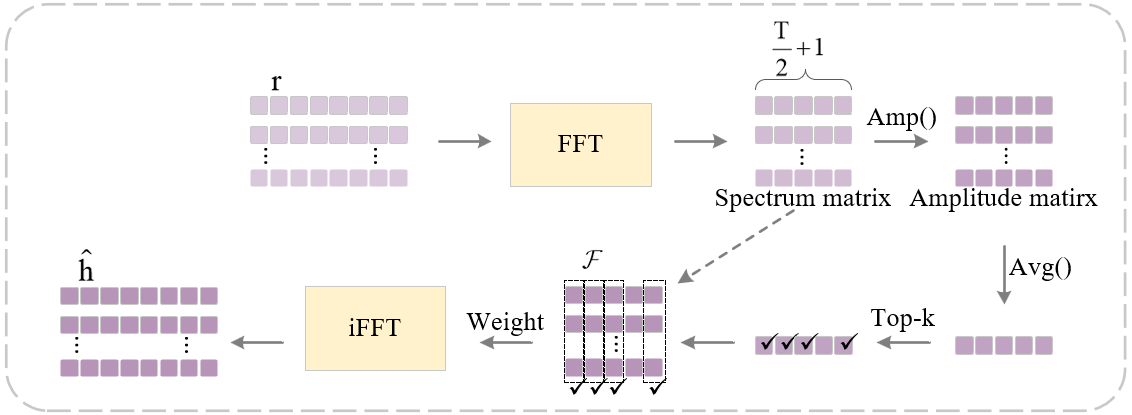}
     \caption{The design of frequency-aware contrastive module.}
     \label{fig:Frequency Domain Feature Extraction Module}
 \end{figure}
 
\subsubsection{Our Design}
%Given the presence of noisy data in time series and the inherent sparsity of the frequency domain\cite{zhou2022fedformer}, we adopt a strategy inspired by the approach used in TimesNET\cite{WuHLZ0L23}. 
The model diagram of FACM is shown in Figure \ref{fig:Frequency Domain Feature Extraction Module}.
Inspired by \cite{WuHLZ0L23}, we aim to capture the most prominent frequencies by retaining only the top \(k\) amplitude values from the representation \(\mathbf{r}\), thereby preserving key spectral information with unnormalized amplitude. Specifically, we begin by applying the FFT to \(\mathbf{r}\), which yields a complex-valued spectrum matrix. Subsequently, the amplitude of this spectrum is computed using the \(\text{Amp}()\) function, followed by averaging over the feature dimension \(K\) to obtain the mean amplitude of each frequency component, denoted as \(\mathbf{A}\). Finally, the \(\text{Top}_k()\) function retains only the top \(k\) frequencies  from \(\mathbf{A}\) , which maximizes spectral focus by discarding lower-amplitude frequencies. This process is formally described as follows:
\begin{equation}
\mathbf{A} = \mathrm{Avg} \left( \mathrm{Amp} \left( \mathrm{FFT}(\mathbf{r}) \right) \right)
\end{equation}
\begin{equation}
\mathcal{F}=\{f_1, \cdots, f_k, 0, \cdots, 0\} = \arg \mathrm{Top}_k \left( \mathbf{A} \right)_{f_* \in \{1, \cdots, \left [ \frac{T}{2}  \right ] \}}
\end{equation}\label{eq:freq}
where \(\mathcal{F} \in \mathbb{R}^{c\times D}\) denotes the set of frequencies with the highest amplitudes for instance \(i\), and $c= \left\lfloor \frac{T}{2}  \right\rfloor +  1$.
This sparse selection method effectively addresses the challenges of the frequency domain’s sparsity and minimizes noise, especially from irrelevant high-frequency components. By focusing on the most significant frequencies, the model captures essential periodic patterns more effectively, filtering out noise and redundant signals.

Next, adaptive weights are assigned to the analyzed frequencies to differentiate the importance of various frequency domains and their corresponding semantic information. These weights are dynamically adjusted to reflect the significance of each frequency component. 
%A learnable weight matrix, $\omega_p$, is applied to create weighted combinations of the selected frequency components. This mechanism adjusts the influence of each frequency according to its importance, emphasizing the contribution of more significant frequencies for specific tasks.
\begin{equation}
\mathbf{\hat{h}}= \mathcal{F}\cdot \omega_p + \beta_p
\end{equation}
Here, $\beta_p$ and $\omega_p$ represent the bias matrix and the learnable weight matrix, calculated as\cite{eldele2024tslanet}.
To maximize the stability and generalizability of the model, we explore diverse initialization schemes or regularization techniques for these adaptive weights. It is ensured that the model remains robust across datasets with varying frequency characteristics, balancing learning stability with performance.

Finally, we apply the iFFT operation to convert the data back to the time domain, followed by a dropout operation to randomly mask certain data points. Further details on the FACM module procedure are provided in the Algorithm \ref{alg:mask_frequency}. 
\begin{equation}
\mathbf{\hat{h}}  = \text{Dropout}(\text{iFFT}(\mathbf{\hat{h}}))
\end{equation}

\begin{algorithm}
\caption{FACM Algorithm}
\textbf{Input:} $\mathbf{r} \in \mathbb{R}^{T \times K}$ \\
\textbf{Parameter:} Masking rate ${\alpha}_m $ \\
\textbf{Output:} Augmented frequency domain feature $\mathbf{\hat{h}}$ 
\begin{algorithmic}[1]
\State Perform FFT transformation on the input representations vector to obtain $\mathcal{F} \in \mathbb{R}^{c \times D }, c= \left\lfloor \frac{T}{2}  \right\rfloor +  1$;
\State Create trainable weight parameters $\omega_p \in \mathbb{R}^{D \times \frac{K}{2}}$ and bias parameters $\beta_p \in \mathbb{R}^{c \times \frac{K}{2}}$;
\State Compute the average amplitude, select the $k$ largest amplitudes, and mask the remaining ones, $k=  c\ast {\alpha}_m$;
\State Assign weights to the selected frequency components using the weight parameters $\omega_p$, then add the bias parameters $\beta_p$;
\State Use the iFFT operation to transform $\mathcal{F}$ back into time-domain representations $\mathbf{\hat{h}}   \in \mathbb{R}^{T \times \frac{K}{2}}$;
\State \textbf{Return} $\mathbf{\hat{h}}$
\end{algorithmic}
\label{alg:mask_frequency}
\end{algorithm}

\subsubsection{Dual Frequency Contrastive Loss} 
To enhance the ability of MFF-FTNet to distinguish between different seasonal and periodic patterns, we incorporate a more granular approach to contrastive learning. Inspired by CoST \cite{woo2022cost}, which demonstrates that frequency components can be uniquely characterized by their amplitude and phase, we define contrastive loss functions, $\mathcal{L}_{\text{amp}}$ and $\mathcal{L}_{\text{phase}}$, to apply contrastive loss specifically on these amplitude and phase aspects. This approach allows the model to better capture and differentiate essential frequency characteristics, thereby improving its capacity to recognize diverse seasonal and periodic patterns.

For the loss functions $\mathcal{L}_{\text{amp}}$ and $\mathcal{L}_{\text{phase}}$, 
we define two augmented versions of the same sample as a positive pair, while different samples within the same batch are considered negative pairs. Each augmented version is derived from distinct augmentation procedure, and then processed separately through the backbone module and the FACM.
Frequency can be uniquely represented by its amplitude and phase vectors, $\acute{f}$ and $\grave{f} \in \mathbb{R}^{c \times D}$, respectively. Using this approach, we compute the losses separately for each positive and negative pair to effectively capture the distinctiveness of the amplitude and phase components.
\begin{equation}
\mathcal{L}_{\text{amp}} = 
\frac{1}{c} 
% \sum_{i=1}^{B}
\sum_{j=1}^{c} 
    % \mathcal{L}_{\text{amp}}^{i,j} = 
    -\log \frac{\exp\left(\acute{f}_{j}^1  \cdot  \acute{f}_{j}^2   \right)}{\exp\left(\acute{f}_{j}^1 \cdot  \acute{f}_{j}^2  \right) + \sum_{k=1}^{c} \exp\left(  \acute{f}_{j}^1 \cdot \acute{f}_{k}^2 \right)}
\end{equation}
\begin{equation}
\mathcal{L}_{\text{phase}} = \frac{1}{c} 
\sum_{j=1}^{c} 
    -\log \frac{\exp\left( \grave{f}_{j}^1  \cdot   \grave{f}_{j}^2   \right)}{\exp\left(  \grave{f}_{j}^1  \cdot  \grave{f}_{j}^2  \right) + \sum_{k=1}^{c} \exp\left(  \grave{f}_{j}^1  \cdot  \grave{f}_{k}^2  \right)}
\end{equation}
where $k \neq j $, $f_{j}^1$ and $f_{j}^2$ denote the two augmented version of the $j$-th frequency sample, 
and $c$ represents the number of frequencies. 
To represent the contrastive loss with the amplitude and phase components, a tunable parameter $\lambda$ is introduced to control the balance between them. This parameter allows flexibility in emphasizing either the amplitude or phase component, depending on the characteristics of the dataset, thus enhancing the model's adaptability across datasets with varying periodic patterns. The contrastive loss function for  FACM is defined as follows:
\begin{equation}
\mathcal{L}_{\text{freq}}=\lambda\mathcal{L}_{\text{amp}}+(1-\lambda)\mathcal{L}_{\text{phase}}
\end{equation}

\subsection{Complementary Time Domain Contrastive Module}
\subsubsection{Motivation}
The Complementary Time Domain Contrastive module (CTCM) is designed to extract essential time-domain features, complementing the insights gained from the FACM. While the FACM effectively captures overarching periodic patterns and high-level structures by focusing on prominent frequency components, the CTCM emphasizes temporal characteristics and localized patterns. This allows it to capture detailed variations and intricate dependencies within the raw time-series data. 
%In traditional time-series analysis, the one-dimensional structure primarily captures changes between consecutive time points, limiting its effectiveness in integrating multi-scale features across the series. To address this limitation, and inspired by \cite{WuHLZ0L23}, we propose a novel time-domain feature extraction module, 
\subsubsection{Our Design}
\begin{figure}
     \centering
     \includegraphics[width = \linewidth]{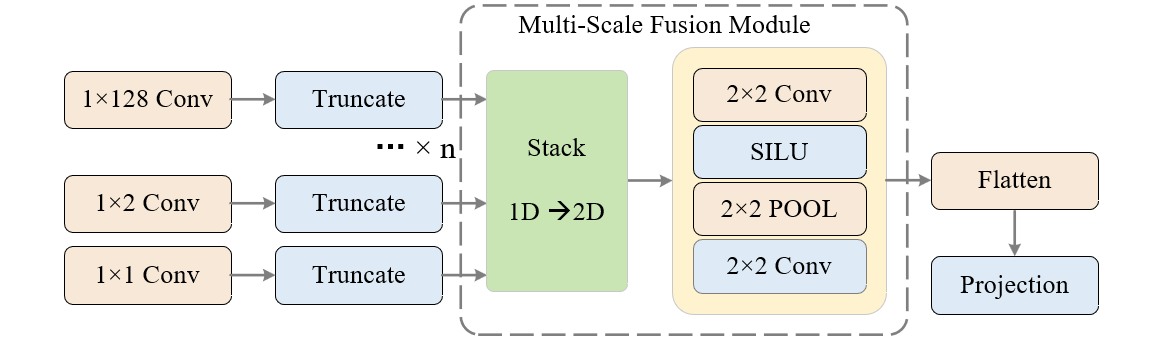}
     \caption{The design of complementary time domain contrastive module.}
     \label{fig:Time Domain Feature Extraction Module}
 \end{figure}

As illustrated in Figure \ref{fig:Time Domain Feature Extraction Module}, the module initially employs multiple parallel 1D convolutional layers with various kernel sizes to  process the input time-series data representation, aiming to simultaneously obtain fine-grained local information and coarse-grained global information. Each kernel size corresponds to a different temporal range, intuitively, smaller kernels capture short-term features, while larger kernels capture patterns over longer time spans.
In our experiments, we investigate the impact of the number of 1D convolutional layers, $n$, on performance. Optimal results are achieved with $n=8$ and kernel sizes of [1,2,4,8,16,32,64,128]. This setup effectively captures the multi-scale, periodic nature of time-series data, which often includes features tied to daily, weekly, or seasonal cycles (e.g., 12-hour or 24-hour intervals).
%Given the latent representation $\mathbf{r}^{i} \in \mathbb{R}^{T \times D}$ from backbone encoder, this representation is processed paralleled through multiple 1D convolutional operations, each with a distinct kernel size. This multi-scale convolutional approach enables the model to capture temporal patterns across different time spans. 
To ensure consistency in output length across these convolutional layers, a truncation operation is applied to the output of each convolution, aligning them for further processing. The process is formulated as follows:
\begin{equation}
\begin{split}
\mathbf{h}_j&= \text{Truncate}(\text{Conv1d}(\mathbf{r}, kernel_j)),
\\ kernel_j &\in [1, 2, 4, 8, 16, 32, 64, 128], j\in\{1,\cdots,n\}\\
\mathbf{h}_{D} &= \text{Stack}([\mathbf{h}_1, \mathbf{h}_2,\dots,\mathbf{h}_n])\in \mathbb{R}^{ K \times n \times T}
\end{split}
\end{equation}
Next, $\mathbf{h}_{D}$ is fed into a Multi-scale Feature Fusion(MSFF) module.
As shown in Figure \ref{fig:Time Domain Feature Extraction Module}, MSFF consists of a sequence of 2D convolution operations, SiLU activation function and average pooling. In particular, 2D convolution operates across both the temporal and feature dimensions, capturing complex dependencies among features of varying time scales.
The activation function (e.g., SiLU) introduces non-linearity, enabling the model to capture more intricate patterns and enrich feature representation.
Average pooling is used to further aggregate the information, reducing dimensionality and preserving essential features while removing redundancies. Therefore, 
MSFF effectively integrates multi-scale features, strengthening the model's global feature representation. This prepares the data for subsequent contrastive learning tasks, providing a comprehensive and unified representation of the time series. Specifically, MSFF module is formulated as:
%In the 2D feature extraction module, the 2D convolutional layer is first applied to fuse the multi-scale feature representations. Afterward, the Silu activation function is utilized, followed by aggregating the extracted information through average pooling.
\begin{equation}
\mathbf{h}_{2D} = \text{Conv2d}(\text{POOL}(\text{SILU}(\text{Conv2d}(\mathbf{h}_{D}))))
\end{equation}
Finally, we apply a flattening operation to convert the two-dimensional representation into a one-dimensional sequence. This transformation simplifies the data structure, making it suitable for subsequent loss function calculations.
\begin{equation}
\mathbf{\tilde{h}} = \text{Projection}(\text{Flatten}(\mathbf{h}_{2D})) 
\end{equation} 
%Here, $\mathbf {\tilde{h}}^{i}$ represents the final extracted feature in the time domain.\\[10pt]
\subsubsection{Time Contrastive Loss}
To achieve a comprehensive and enriched representation of the time series, we concatenate the feature representations extracted from the time and frequency domains along the feature dimension. This combined representation is then passed through a linear layer to project it into a shared latent feature space.
\begin{equation}
\mathbf{h} = \text{Liner}(\text{Concat}(\mathbf{\tilde{h}},\mathbf{\hat{h}})) 
\end{equation}
Given $\mathbf{x}_t$, the values of input $\textbf{X}$ at time step $t$, we obtain two corresponding feature representations: the first is $r_t$, the feature vector from the backbone, and the second is $h_t$, which combines representation results from both time and frequency domains. To learn discriminative temporal representations, we treat the pair $r_t$ and $h_t$, derived from the same input $x_t$ within the same batch, as a positive pair. Representations obtained from different inputs within the same batch are treated as negative pairs. The loss function is defined as: 
\begin{equation}
    \mathcal{L}_{\text{time}} =  \sum_{t=1}^{T} -\log \frac{\exp(r_t \cdot h_t)}{\exp(r_t \cdot h_t) + \sum_{t'=1, t'\neq t}^{T} ( \exp(r_t \cdot h_{t'}) )}
\end{equation}

\subsection{Jointly Optimized Training.}
The model is optimized by combining frequency domain loss and feature loss, resulting in the overall loss function as follows. 
\begin{equation}
    \mathcal{L}_{\text{total}} = \gamma_1 \mathcal{L}_{\text{time}} + \gamma_2 \mathcal{L}_{\text{freq}}
\end{equation}
where two loss weight parameters, $\gamma_1$ and $\gamma_2$, are introduced to balance the contributions of these two losses. They control the emphasis on frequency domain information and feature-based information during the model's learning process. This flexibility enables the model to adapt to different time series characteristics and improve its robustness and accuracy across diverse datasets.

\section{EXPERIMENTS}
\subsection{Datasets.} 
We conduct experiments on five publicly available real-world datasets: ETTh1, ETTh2, ETTm1, Weather\cite{woo2022cost}, and ETTm2\cite{zheng2023simts}. Detailed dataset information is provided in Table \ref{table:datasets}. The ETT (Electricity Transformer Temperature) dataset comprises two hourly-level datasets (ETTh) and two 15-minute-level dataset (ETTm), each containing six power load features, with “oil temperature” used as the target variable for univariate forecasting. Similarly, the Weather dataset includes hourly-level climate data with 11 features collected from approximately 1,600 locations across the U.S., with the “wet bulb” temperature selected as the univariate forecasting target. Following previous work, we split each dataset into training, validation, and test sets in a 6:2:2 ratio.

\setlength{\extrarowheight}{2pt}
\begin{table}
    \caption{Detailed Statistics of the Datasets.}
    \centering
    \begin{tabular}{llllll}
        \hline
        Dataset & Samples & Features & Train & Valid & Test  \\ \hline
        ETTh1 & 17,420 & 7 & 8640 & 2880 & 2880  \\ 
        ETTh2 & 17,420 & 7 & 8640 & 2880 &2880  \\ 
        ETTm1 & 69,680 & 7 & 34560 & 11520 &11520  \\
        ETTm2 & 69,680 & 7 & 34560 & 11520 &11520  \\ \hline
        WTH & 35,064 & 12 & 21038 & 7013 & 7013  \\ \hline
    \end{tabular}
    \label{table:datasets}
\end{table}
\subsection{Baselines}
We select five state-of-the-art baseline methods for comparison with MFF-FTNet. These methods are highly representative and remain relevant benchmarks, demonstrating excellent predictive performance. Descriptions and implementations of these baselines are provided below.\\
\indent \textbf{CoST} \cite{woo2022cost}: CoST applies season-trend disentanglement by using the Fourier Transform to independently learn seasonal and trend components. The final representation for forecasting is constructed by combining these separate seasonal and trend representations.

\indent \textbf{TS2Vec} \cite{yue2022ts2vec}: TS2Vec is a method for embedding time series data into fixed-length vectors. It interprets time series as spatiotemporal graphs, employing multi-level convolutions and pooling, along with image processing techniques, to generate compact vector representations.

\indent \textbf{TNC} \cite{tonekaboni2021unsupervised}: TNC uses convolutional neural networks with Fourier transformations in both the time and frequency domains to enhance time series prediction. By incorporating time-frequency features, TNC identifies time series anomalies from a multi-perspective view.

\indent \textbf{Informer} \cite{zhou2021informer}: Informer is a transformer-based model specifically designed for long-sequence time series forecasting. It includes three core components: a ProbSparse self-attention mechanism, self-attention distillation, and a generative-style decoder.

\indent \textbf{TCN} \cite{bai2018empirical}: It leverages CNNs for sequential data processing, replacing loops with convolution operations to handle time series data. 
%TCN uses stacked convolutional layers to extend the receptive field, enabling it to capture long-range dependencies effectively.
% \indent InfoTS\cite{luo2023time}: InfoTS introduces an information-aware data augmentation criterion for time series, focusing on high diversity and fidelity. It approximates this using mutual information and cross-entropy estimation, with a meta-learner network adaptively selecting the best augmentation for contrastive representation learning.\\
% \indent AutoTCL\cite{zheng2024parametric}: AutoTCL can be adaptively applied to support time series representation learning. The proposed method is encoder-agnostic, allowing it to seamlessly integrate with different backbone encoders.\\
% \indent T-Rep\cite{fraikin2023t}: T-Rep learns temporal vector embeddings along with its feature extractor to capture temporal features from the signal, such as trends, seasonality, or distribution shifts. These temporal embeddings are utilized in surrogate tasks to incorporate smooth and fine-grained temporal dependencies into the representations, enhancing robustness against missing data.\\

%\indent The multivariate and univariate prediction results for these baseline methods on the ETTm2 dataset are sourced from SimTS \cite{zheng2023simts}, while the AutoTCL results on ETTm2 are derived from our transfer learning experiments.

\subsection{Experimental Setup}
All experiments were implemented in PyTorch and conducted on a system equipped with an Intel Core i5-13500HX @ 2.50GHz processor and an NVIDIA GeForce GTX 4060 (8 GB) GPU. Prediction horizons ($L$) were tailored to the characteristics of each dataset: for the ETTh1, ETTh2, and Weather datasets, $L$ was set to \{24,48,168,336,720\}; for the ETTm1 and ETTm2 datasets, it was set to 
\{24,48,96,288,672\}. Due to significant differences in time series characteristics across domains, model hyperparameters were adjusted separately for multivariate and univariate forecasting tasks, with detailed settings provided in Table \ref{table: hyperparameter settings}. In both tasks, the batch size was set to 128, training spanned 600 epochs, and the backbone encoder’s output dimension was 320. The output dimensions for FACM and CTCM were both set to 160.
% , while the intermediate vector in CTCM’s Projection block had a dimension of 96. 
This experimental setup ensured that the model was robust and adaptable to a variety of time series forecasting scenarios.

\setlength{\extrarowheight}{2pt}
\begin{table}
    \caption{Hyperparameter settings for different datasets.}
    \centering
    \begin{tabular}{clcccc}
    \hline
    Dataset & Prediction task & $D^ {\prime\prime}$  & $D^ \prime$ & $L^ \prime$ & W\\ \hline
    \multirow{2}{*}{ETT} & Multivariate  & 96  & 32 & 8 & 1e-4 \\
                         & Univariate    & 48  & 96 & 10 & 1e-5 \\ \hline
    \multirow{2}{*}{WTH} & Multivariate  & 96  & 64 & 8 & 1e-4 \\
                         & Univariate    & 96  & 64 & 8 & 1e-4 \\ \hline
    \end{tabular}
    \label{table: hyperparameter settings}
\end{table}

\setlength{\extrarowheight}{2pt}
\begin{table*}
    \caption{Comparison with baselines for multivariate time series forecasting. Bold represents the best performance.}
    \centering
    \begin{tabular}{cccccccccccccc}
        \hline
        \multicolumn{2}{l}{\textbf{Methods}} & \multicolumn{2}{c}{\textbf{MFF-FTNet}} & \multicolumn{2}{c}{\textbf{CoST}} & \multicolumn{2}{c}{\textbf{TS2Vec}} & \multicolumn{2}{c}{\textbf{TNC}} & \multicolumn{2}{c}{\textbf{Informer}} & \multicolumn{2}{c}{\textbf{TCN}}\\ \cline{3-14}

        % \multicolumn{2}{l}{\textbf{Accepted by}} & \multicolumn{2}{c}{\textbf{ours}} & \multicolumn{2}{c}{\textbf{ICLR 2024}}& \multicolumn{2}{c}{\textbf{ICLR 2022}} & \multicolumn{2}{c}{\textbf{ AAAI 2022}} & \multicolumn{2}{c}{\textbf{ICLR 2021}} & \multicolumn{2}{c}{\textbf{AAAI 2023}} & \multicolumn{2}{c}{\textbf{AAAI 2021}} & \multicolumn{2}{c}{\textbf{TCN}}\\ \cline{3-18}
        
        \multicolumn{2}{l}{\textbf{Metrics}} & \textbf{MSE} & \textbf{MAE} & \textbf{MSE} & \textbf{MAE} & \textbf{MSE} & \textbf{MAE}& \textbf{MSE} & \textbf{MAE}& \textbf{MSE} & \textbf{MAE} & \textbf{MSE} & \textbf{MAE}\\ \hline
        \multirow{5}{*}{\rotatebox{90}{\textbf{ETTh1}}} & 24 & \textbf{0.382} & \textbf{0.425}  & 0.386 & 0.429 & 0.590 & 0.531 & 0.708 & 0.592 & 0.577 & 0.549 & 0.583 & 0.547 \\ 
                                               & 48 & \textbf{0.423} & \textbf{0.454}  & 0.437 & 0.464 & 0.624 & 0.555 & 0.749 & 0.619 &  0.685 & 0.625 & 0.670 & 0.606\\ 
                                               & 168 & \textbf{0.621} & \textbf{0.573}  & 0.643 & 0.582 & 0.762 & 0.639 & 0.884 & 0.699 &  0.931 & 0.752 & 0.811 & 0.680\\ 
                                               & 336& \textbf{0.775} & \textbf{0.664}  & 0.812 & 0.679 & 0.931 & 0.728 & 1.020 & 0.768 &  1.128 & 0.873 & 1.132 & 0.815\\ 
                                               & 720& \textbf{0.880} & \textbf{0.737}  & 0.970 & 0.771 & 1.063 & 0.799 & 1.157 & 0.830 &  1.215 & 0.896 & 1.165 & 0.813\\ \hline
        \multirow{5}{*}{\rotatebox{90}{\textbf{ETTh2}}} & 24 & \textbf{0.336} & \textbf{0.425}   & 0.447 & 0.502 & 0.423 & 0.489 & 0.612 & 0.595  & 0.720 & 0.665 & 0.935 & 0.754\\ 
                                               & 48 & \textbf{0.563} & \textbf{0.554}  & 0.699 & 0.637 & 0.619 & 0.605 & 0.840 & 0.716 & 1.457 & 1.001 & 1.300 & 0.911\\ 
                                               & 168 & \textbf{1.378} & \textbf{0.913}  & 1.549 & 0.982 & 1.845 & 1.074 & 2.359 & 1.213 &  3.489 & 1.515 & 4.017 & 1.579\\ 
                                               & 336& \textbf{1.606} & \textbf{0.990}   & 1.749 & 1.042 & 2.194 & 1.197 & 2.782 & 1.349 & 2.723 & 1.340 & 3.460 & 1.456\\ 
                                               & 720& \textbf{1.925} & 1.101  & 1.971 & \textbf{1.092} & 2.636 & 1.370 & 2.753 & 1.394 & 3.467 & 1.473 & 3.106 & 1.381\\ \hline
        \multirow{5}{*}{\rotatebox{90}{\textbf{ETTm1}}} & 24 & \textbf{0.241} & \textbf{0.324}  & 0.246 & 0.329 & 0.453 & 0.444 & 0.522 & 0.472 & 0.323 & 0.369 & 0.363 & 0.397\\ 
                                               & 48 & \textbf{0.321} & \textbf{0.380} & 0.331 & 0.386 & 0.592 & 0.521 & 0.695 & 0.567 & 0.494 & 0.503 & 0.542 & 0.508\\ 
                                               & 96 & \textbf{0.370} & \textbf{0.414}  & 0.378 & 0.419 & 0.635& 0.554 & 0.732 & 0.595 & 0.678 & 0.614 & 0.666 & 0.578\\ 
                                               & 288 & \textbf{0.454} & \textbf{0.474}   & 0.472 & 0.486 & 0.693 & 0.597 & 0.818 & 0.649 &1.056 & 0.786 & 0.991 & 0.735\\ 
                                               & 672& \textbf{0.610} & \textbf{0.569}  & 0.620 & 0.574 & 0.782 & 0.653 & 0.932 & 0.712 &  1.192 & 0.926 & 1.032 & 0.756\\ \hline
        \multirow{5}{*}{\rotatebox{90}{\textbf{ETTm2}}} & 24 & 0.123 & 0.247  & \textbf{0.122} & \textbf{0.244} & 0.180 & 0.293 & 0.185 & 0.297 & 0.173 & 0.301 & 0.180 & 0.324\\ 
                                               & 48 & \textbf{0.180} & 0.307   & 0.183 & \textbf{0.305} & 0.244 & 0.350 & 0.264 & 0.360 & 0.303 & 0.409 & 0.204 & 0.327\\ 
                                               & 96 & \textbf{0.287} & \textbf{0.391}   & 0.294 & 0.394 & 0.360 & 0.427 & 0.389 & 0.458 & 0.365 & 0.453 & 3.041 & 1.330\\ 
                                               & 288 & \textbf{0.681} & \textbf{0.623}  & 0.723 & 0.652 & 0.723 & 0.639 & 0.920 & 0.788 &  1.047 & 0.804 & 3.162 & 1.337\\ 
                                               & 672& \textbf{1.437} & \textbf{0.928}  & 1.899 & 1.073 & 1.753 & 1.007 & 2.164 & 1.135 & 3.126 & 1.302 & 3.624 & 1.484\\ \hline
        \multirow{5}{*}{\rotatebox{90}{\textbf{WTH}}} & 24 & \textbf{0.297} & \textbf{0.359}  & 0.298 & 0.360 & 0.307 & 0.363 & 0.320 & 0.373 & 0.335 & 0.381 & 0.321 & 0.367\\ 
                                               & 48 & \textbf{0.358} & \textbf{0.410}  & 0.359 & 0.411 & 0.374 & 0.418 & 0.380 & 0.421 & 0.395 & 0.459 & 0.386 & 0.423\\ 
                                               & 168 & \textbf{0.462} & \textbf{0.490}  & 0.464 & 0.491 & 0.491 & 0.506 & 0.479 & 0.495 &  0.608 & 0.567 & 0.491 & 0.501\\ 
                                               & 336& \textbf{0.497} & 0.517  &0.497 & 0.517 & 0.525 & 0.530 & 0.505 & 0.514 & 0.702 & 0.620 & 0.502 & \textbf{0.507}\\ 
                                               & 720& 0.534 & 0.542 & 0.533 & 0.542 & 0.556 & 0.552 & 0.519 & 0.525 &  0.831 & 0.731 & \textbf{0.498} & \textbf{0.508}\\ \hline
        \multicolumn{2}{c}{\textbf{Avg.}}  & \textbf{0.630} & \textbf{0.552}  & 0.683 & 0.575 & 0.814 & 0.633 & 0.940 & 0.686 & 1.070 & 0.754 & 1.063 & 0.712 \\ \hline
    \end{tabular}
    \label{table:comparison multivariate time series forecasting}
\end{table*}

\setlength{\extrarowheight}{2pt}
\begin{table*}[htbp]
    \caption{Comparison with baselines for univariate time series forecasting. Bold represents the best performance.}
    \centering
    \begin{tabular}{cccccccccccccc}
        \hline
        \multicolumn{2}{l}{\textbf{Methods}} & \multicolumn{2}{c}{\textbf{MFF-FTNet}} & \multicolumn{2}{c}{\textbf{CoST}} & \multicolumn{2}{c}{\textbf{TS2Vec}} & \multicolumn{2}{c}{\textbf{TNC}} & \multicolumn{2}{c}{\textbf{Informer}} & \multicolumn{2}{c}{\textbf{TCN}}\\ \cline{3-14}
        
        \multicolumn{2}{l}{\textbf{Metrics}} & \textbf{MSE} & \textbf{MAE} & \textbf{MSE} & \textbf{MAE} & \textbf{MSE} & \textbf{MAE}& \textbf{MSE} & \textbf{MAE} & \textbf{MSE} & \textbf{MAE}& \textbf{MSE} & \textbf{MAE}\\ \hline
\multirow{5}{*}{\rotatebox{90}{\textbf{ETTh1}}} & 24 & \textbf{0.036} & \textbf{0.145} & 0.040 & 0.152 & 0.039 & 0.151 & 0.057 & 0.184 & 0.098 & 0.147 & 0.104 & 0.254  \\ 
                                               & 48 & \textbf{0.055} & \textbf{0.179} & 0.060 & 0.186 & 0.062 & 0.189 & 0.094 & 0.239 & 0.158 & 0.319 & 0.206 & 0.366 \\ 
                                               & 168 & \textbf{0.090} & \textbf{0.228} & 0.097 & 0.236 & 0.142 & 0.291 & 0.171 & 0.329 & 0.183 & 0.346 & 0.462 & 0.586 \\ 
                                               & 336& \textbf{0.102} & \textbf{0.243} & 0.112 & 0.258 & 0.160 & 0.316 & 0.179 & 0.345 & 0.222 & 0.387 & 0.422 & 0.564 \\ 
                                               & 720& \textbf{0.134} & \textbf{0.288} & 0.148 & 0.306 & 0.179 & 0.345 & 0.235 & 0.408 & 0.269 & 0.435 & 0.438 & 0.578 \\ \hline
\multirow{5}{*}{\rotatebox{90}{\textbf{ETTh2}}} & 24 & \textbf{0.080} & \textbf{0.207} & 0.079 & 0.207 & 0.091 & 0.230 & 0.097 & 0.238 & 0.093 & 0.240 & 0.109 & 0.251 \\ 
                                               & 48 & \textbf{0.114} & \textbf{0.255} & 0.118 & 0.259 & 0.124 & 0.274 & 0.131 & 0.281 & 0.155 & 0.314 & 0.147 & 0.302 \\ 
                                               & 168 & \textbf{0.172} & \textbf{0.325} & 0.189 & 0.339 & 0.198 & 0.355 & 0.197 & 0.354 & 0.232 & 0.389 & 0.209 & 0.366 \\ 
                                               & 336& \textbf{0.200} & \textbf{0.356} & 0.206 & 0.360 & 0.205 & 0.364 & 0.207 & 0.366 & 0.263 & 0.417 & 0.237 & 0.391  \\ 
                                               & 720& 0.212 & 0.372 & 0.214 & 0.371 & 0.208 & 0.371 & 0.207 & 0.370 & 0.277 & 0.431 & \textbf{0.200} & \textbf{0.367} \\ \hline
\multirow{5}{*}{\rotatebox{90}{\textbf{ETTm1}}} & 24 & \textbf{0.013} & \textbf{0.084} & 0.015 & 0.088 & 0.016 & 0.093 & 0.019 & 0.103 & 0.030 & 0.137 & 0.027 & 0.127\\ 
                                               & 48 & \textbf{0.024} & \textbf{0.114} & 0.025 & 0.117 & 0.028 & 0.126 & 0.045 & 0.162 & 0.069 & 0.203 & 0.040 & 0.154 \\ 
                                               & 96 & \textbf{0.038} & \textbf{0.146} & 0.038 & 0.147 & 0.045 & 0.162 & 0.054 & 0.178 & 0.194 & 0.372 & 0.097 & 0.246 \\ 
                                               & 288 & \textbf{0.072} & \textbf{0.202} & 0.077 & 0.209 & 0.095 & 0.235 & 0.142 & 0.290 & 0.401 & 0.544 & 0.305 & 0.455\\ 
                                               & 672& \textbf{0.102} & \textbf{0.240} & 0.113 & 0.257 & 0.142 & 0.290 & 0.136 & 0.290 & 0.277 & 0.431 & 0.200 & 0.367\\ \hline
\multirow{5}{*}{\rotatebox{90}{\textbf{ETTm2}}} & 24 & \textbf{0.024} & \textbf{0.106} & 0.027 & 0.112 & 0.038 & 0.139 & 0.045 & 0.151 & 0.036 & 0.141 & 0.048 & 0.153 \\ 
                                               & 48 & \textbf{0.049} & \textbf{0.157} & 0.054 & 0.159 & 0.069 & 0.194 & 0.080 & 0.201 & 0.069 & 0.200 & 0.063 & 0.191 \\ 
                                               & 96 & \textbf{0.073} & \textbf{0.196} & 0.072 & 0.196 & 0.089 & 0.225 & 0.094 & 0.229 & 0.095 & 0.240 & 0.129 & 0.265\\ 
                                               & 288 & \textbf{0.139} & \textbf{0.274} & 0.153 & 0.307 & 0.161 & 0.306 & 0.155 & 0.309 & 0.211 & 0.367 & 0.208 & 0.352\\ 
                                               & 672& \textbf{0.179} & \textbf{0.317} & 0.183 & 0.329 & 0.201 & 0.351 & 0.197 & 0.352 & 0.267 & 0.417 & 0.222 & 0.377\\ \hline
\multirow{5}{*}{\rotatebox{90}{\textbf{WTH}}} & 24 & \textbf{0.096} & \textbf{0.212} & 0.096 & 0.213 & 0.096 & 0.215 & 0.102 & 0.221 & 0.117 & 0.251 & 0.109 & 0.217\\ 
                                               & 48 & \textbf{0.138} & \textbf{0.263} & 0.138 & 0.262 & 0.140 & 0.264 & 0.139 & 0.264 & 0.178 & 0.318 & 0.143 & 0.269\\ 
                                               & 168 & 0.217 & 0.342 & 0.207 & 0.334 & 0.207 & 0.335 & 0.198 & 0.328 & 0.266 & 0.398 & \textbf{0.188} & \textbf{0.319}\\ 
                                               & 336& 0.243 & 0.365 & 0.230 & 0.356 & 0.231 & 0.360 & 0.215 & 0.347 & 0.197 & 0.416 & \textbf{0.192} & \textbf{0.320}\\ 
                                               & 720& 0.252 & 0.377 & 0.242 & 0.370 & 0.233 & 0.365 & 0.219 & 0.353 & 0.359 & 0.466 & \textbf{0.198} & \textbf{0.329}\\ \hline
        \multicolumn{2}{c}{\textbf{Avg.}}  & \textbf{0.114} & \textbf{0.239}& 0.117 & 0.245 & 0.128 & 0.260 & 0.136 & 0.275 & 0.188 & 0.333 & 0.188 & 0.326 \\ \hline
    \end{tabular}
    \label{table:comparison univariate time series forecasting}
\end{table*}

\subsection{Comparison Results}
Tables \ref{table:comparison multivariate time series forecasting} and \ref{table:comparison univariate time series forecasting} summarize the performance of MFF-FTNet and baseline models on five datasets for both multivariate and univariate forecasting across various forecasting horizons. Across all datasets and horizons, MFF-FTNet consistently outperforms the baseline models in both forecasting tasks.

In the \textbf{multivariate time series} forecasting task, MFF-FTNet reduces the average MSE by 7.7\% and the average MAE by 4.0\%, delivering substantial improvements over other models. Notably, MFF-FTNet achieves an 8.9\% reduction in MSE and a 4.4\% reduction in MAE across the four ETT datasets compared to CoST, the top-performing baseline. This outcome underscores the advantages of self-supervised contrastive learning models like MFF-FTNet, CoST, and TS2Vec, which generally surpass traditional methods in predictive accuracy. Frequency-based models, such as MFF-FTNet and CoST, perform particularly well because time series data often contain essential periodic features that frequency analysis can effectively capture. Unlike traditional methods that rely solely on time-domain information, frequency-based models reveal underlying periodic patterns crucial for accurate forecasting. Specifically, MFF-FTNet combines frequency-based periodic information with local and global time-domain trends, enabling a more comprehensive understanding of complex time series data.

% For short-term forecasting at prediction lengths of \{24,48\}, MFF-FTNet shows incremental improvements, with particularly notable gains over CoST as the prediction length increases. This advantage is due to MFF-FTNet’s capacity to capture long-range dependencies effectively, modeling interactions across distant time points, and offering a more comprehensive representation of complex relationships through multi-scale feature modeling. For instance, on the ETTh2 dataset, MFF-FTNet reduces the average MSE by 9.4\% and the average MAE by 6.3\% compared to CoST.

In the \textbf{univariate time series} forecasting, MFF-FTNet achieves the best average results for both evaluation metrics across all datasets. In comparison to CoST, MFF-FTNet reduces the MSE by 2.5\% and the MAE by 2.4\% on average across the five datasets. Additionally, for the ETT series datasets, MFF-FTNet demonstrates significant gains, with a 5.9\% decrease in MSE and a 3.9\% decrease in MAE, highlighting its strong adaptability and accuracy in modeling the patterns within ETT data.

%Overall, these experimental results underscore the ability of MFF-FTNet to model long-range dependencies and capture rich feature information across both time and frequency domains. By integrating information from these domains, MFF-FTNet achieves enhanced pattern recognition and feature extraction for complex time series data, resulting in more accurate and robust forecasting performance.

\subsection{Ablation Analysis}
To verify the effectiveness of the proposed components in MFF-FTNet, we compare the full model with several of its variants on the ETTh1 and ETTm2 datasets. The experimental results are presented in Table \ref{table:Ablation results}. The variants considered include: (1) removing data augmentation \textbf{(w/o DA)}; (2) removing the Frequency-Aware Contrastive Module (FACM) \textbf{(w/o FM)}; (3) removing the Complementary Time Domain Contrastive Module (CTCM) \textbf{(w/o CM)}; and (4) replacing the SiLU activation function with the GELU activation function in the backbone encoder \textbf{(w/o Si)}. Additionally, we explore combinations of these variations. The full model serves as the baseline, and we progressively remove one or more components to analyze their individual contributions in the ablation study.
\setlength{\extrarowheight}{2pt} % 增加每一行的额外高度
\begin{table}[htbp]
    \caption{Ablation results.}
    \centering
    \begin{tabular}{lllllll}
        \hline
        \multirow{2}{*}{\textbf{Model Variants}} & \multirow{2}{*}{\phantom{XXX}} &\multicolumn{2}{c}{\textbf{ETTh1}} & \multicolumn{2}{c}{\textbf{ETTm2}} \\ 
         & & \textbf{MSE} & \textbf{MAE} & \textbf{MSE} & \textbf{MAE} \\ \hline
         MFF-FTNet & & 0.616 & 0.571 & 0.542 & 0.499 \\ \hline
        w/o DA & & 0.621 & 0.573 & 0.543 & 0.501 \\ 
        w/o FM & & 0.624 & 0.573 & 0.544 & 0.503 \\ 
        w/o CM & & 0.638 & 0.579 & 0.543 & 0.501 \\ 
        w/o DA + FM & & 0.628 & 0.575 & 0.546 & 0.503 \\ 
        w/o DA + CM & & 0.639 & 0.579 & 0.543 & 0.501 \\ 
        w/o CM + FM & & 0.629 & 0.576 & 0.545 & 0.505 \\ 
        % w/o DA + CM + FM & & 0.631 & 0.577 & 0.546 & 0.505 \\  
        w/o Si & & 0.621 & 0.574 & 0.552 & 0.511 \\
        \hline
    \end{tabular}
    \label{table:Ablation results}
\end{table}
From the experimental results, we can draw several key conclusions regarding the importance of each module in MFF-FTNet: (1) The removal of data augmentation leads to a noticeable drop in prediction performance. This highlights the importance of data augmentation, which generates diverse training samples, thus enhancing the model’s robustness and generalization ability. (2)
When the FACM (w/o FM) is omitted, there is a significant decline in performance. This demonstrates the critical role of the frequency feature extraction, which mitigates noise caused by high-frequency components and addresses sparsity in the frequency domain. 
(3) Excluding CTCM (w/o CM) results in a substantial decrease in prediction accuracy. This indicates that the model’s ability to capture rich temporal and frequency domain features is severely compromised without the integration of both time and frequency domain contrastive learning. (4) Replacing the SiLU activation function with GELU in the backbone encoder (w/o Si) leads to a noticeable reduction in prediction performance. This confirms that the SiLU activation function is more suitable for time series analysis tasks, as it provides smoother gradients and enhances the model's convergence and robustness.

\subsection{Convolution Kernels in CTCM Study}
\begin{figure}
    \centering
    % 第一行的两张图片
    \begin{minipage}{0.49\columnwidth}
        \centering
        \includegraphics[width=\linewidth]{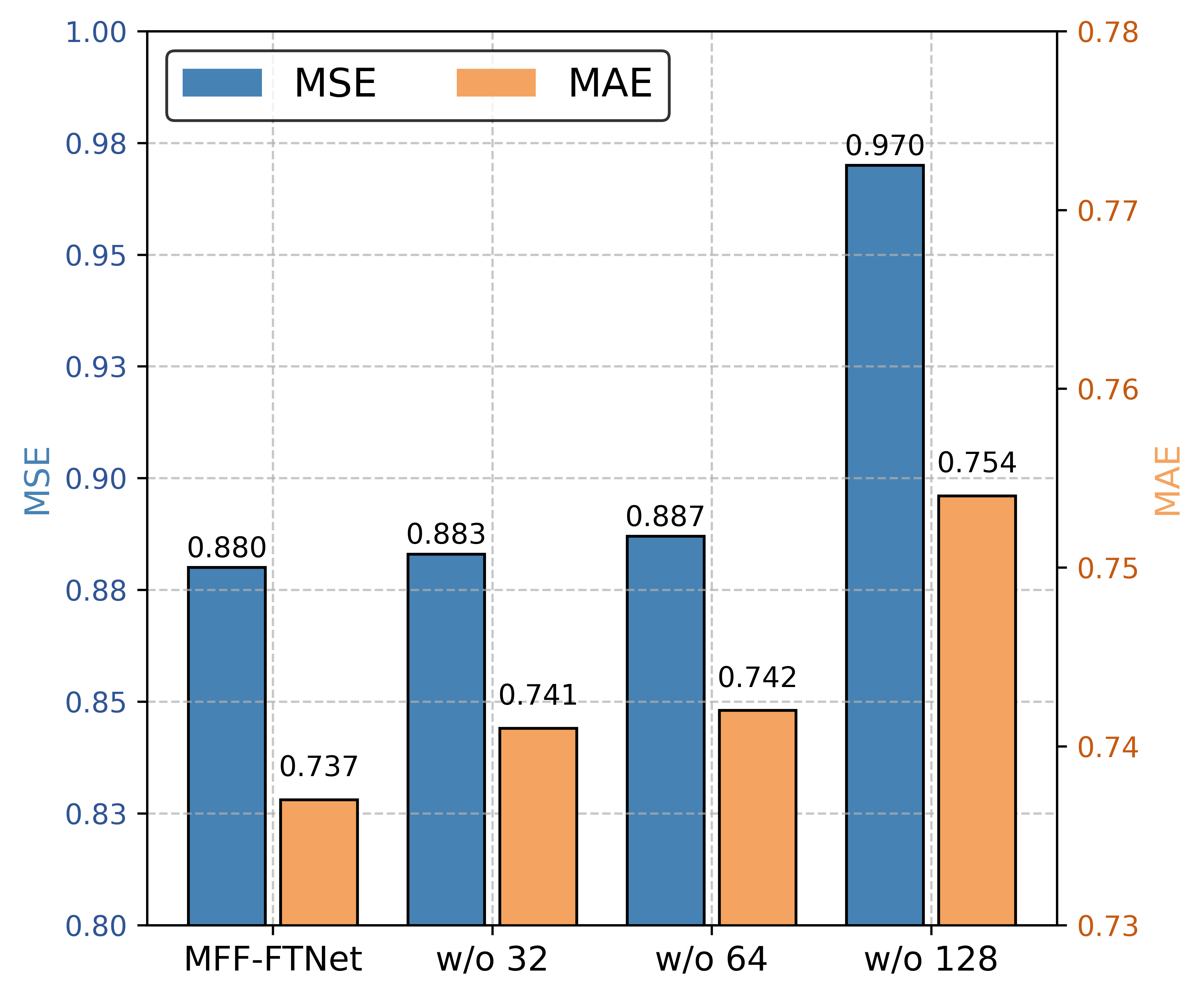}
        \caption*{ETTh1-720}
    \end{minipage}
    \hfill
    \begin{minipage}{0.49\columnwidth}
        \centering
        \includegraphics[width=\linewidth]{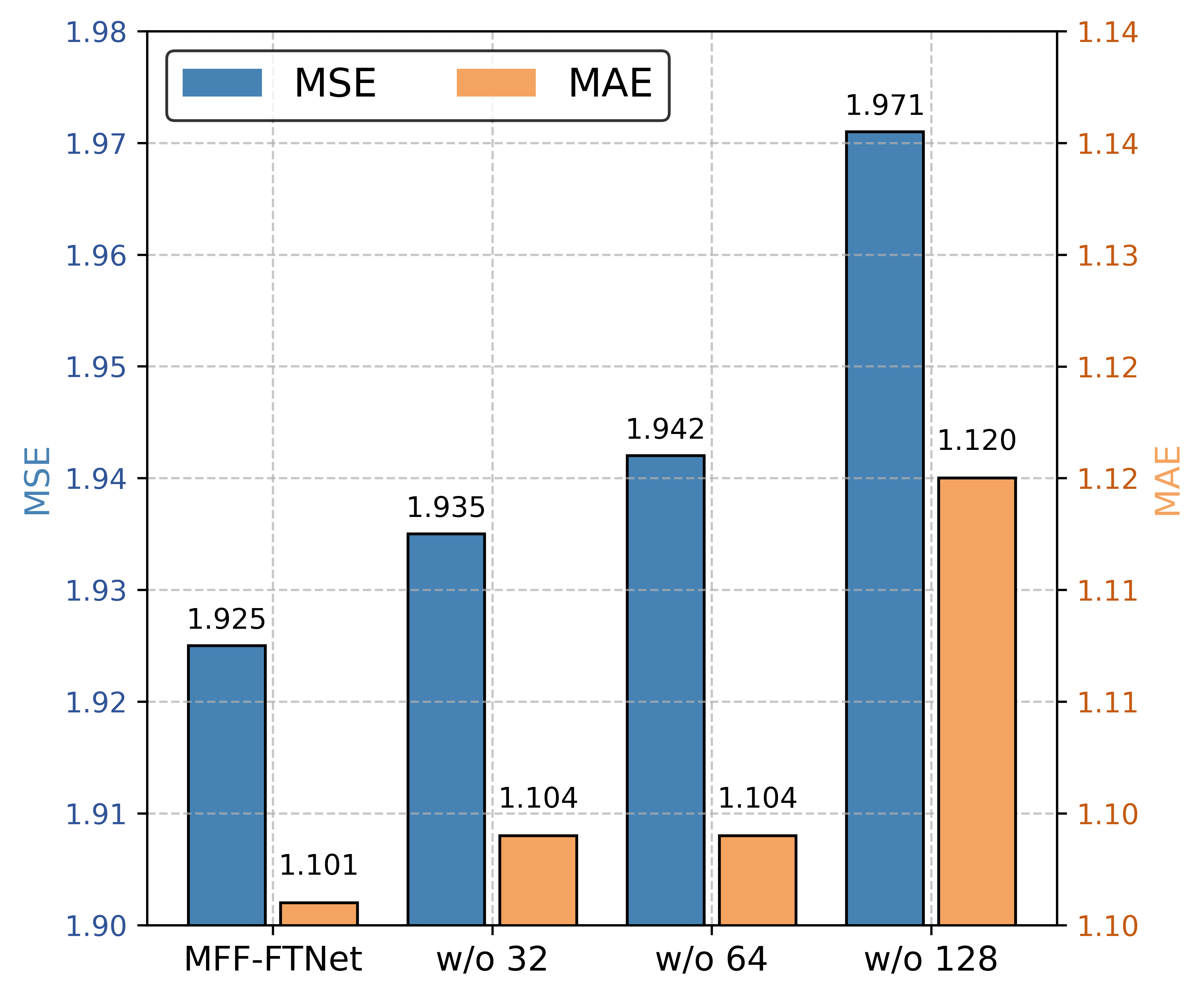}
        \caption*{ETTh2-720}
    \end{minipage}
    \caption{The study on the impact of convolution kernel parameters on prediction performance.}
    \label{fig:kernels}
\end{figure}

To assess the importance of different convolution kernel sizes in CTCM for capturing both local features and long-term dependencies, we conducted multivariate prediction experiments on the ETTh1 and ETTh2 datasets with a prediction length of 720. In these experiments, we progressively removed convolution layers corresponding to small (w/o 32), medium (w/o 64), and large kernel sizes (w/o 128) to analyze the impact on forecasting accuracy.
As shown in Figure \ref{fig:kernels}, removing the larger kernel size of 128 resulted in the most significant decline in prediction performance, indicating that large kernels play a crucial role in capturing global trends and long-term dependencies. These larger kernels allow the model to aggregate information across extended time spans, which is essential for accurate long-horizon forecasting. 

The medium-sized kernel (w/o 64) also contributed notably to model performance, with its removal causing a moderate decline. This suggests that medium-sized kernels are important for capturing mid-range dependencies that link local and global features. 
In contrast, removing the smallest kernel (w/o 32) led to a relatively minor impact on long-term forecasting performance, indicating that while small kernels contribute to capturing fine-grained, local details, they are less critical for extended prediction tasks. 
Overall, these results highlight the effectiveness of employing a range of kernel sizes within CTCM.

\setlength{\extrarowheight}{2pt}
\begin{table*}[htbp]
    \caption{Comparison of transfer learning results.}
    \centering
    \begin{tabular}{ c c| c c c c | c c c c c c}
        \hline
        \multicolumn{2}{c|}{ \multirow{2}{*}{Models}}& \multicolumn{4}{c|}{MFF-FTNet} &\multicolumn{2}{c}{\multirow{2}{*}{CoST}} &\multicolumn{2}{c}{\multirow{2}{*}{AutoTCL}} &\multicolumn{2}{c}{\multirow{2}{*}{T-Rep}}\\ \cline{3-6}
        & & \multicolumn{2}{c|}{SSL.} & \multicolumn{2}{c|}{Fine-tuning} & \multicolumn{2}{c}{ICLR 2022} & \multicolumn{2}{c}{ICLR 2024}& \multicolumn{2}{c}{ICLR 2024} \\ \hline
        \multicolumn{2}{c|}{Metric} & MSE & MAE & MSE & MAE & MSE & MAE & MSE & MAE& MSE & MAE\\ \hline
        \multirow{5}{*}{\textbf{WTH} $\mathbf{\rightarrow}$ \textbf{ETTh1}} & 24 & 0.382 & 0.425 & \textbf{0.375}  & \textbf{0.420}  & 0.386 & 0.429 &  0.389 &  0.439 & 0.511  &0.496   \\
                                                        & 48 & 0.423 & 0.454 & \textbf{0.421}  & \textbf{0.451}  & 0.437 & 0.464& 0.447  &  0.477 &  0.546 &  0.524  \\
                                                        & 168 & 0.621 & 0.573 & \textbf{0.609}  & \textbf{0.564}  & 0.643 & 0.582&  0.615 &  0.574  & 0.759  &  0.649 \\
                                                        & 336 & 0.775 & 0.664 & \textbf{0.763}  & \textbf{0.654}   & 0.812 & 0.679&  0.802 & 0.671  &  0.936 & 0.742   \\
                                                        & 720 & \textbf{0.880} & 0.737 & 0.888 &  \textbf{0.735}  & 0.970 & 0.771& 1.028  &  0.789 & 1.061  &  0.813  \\ \hline
        
        \multirow{5}{*}{\textbf{WTH} $\mathbf{\rightarrow}$\textbf{ETTm1}} & 24 & 0.241 & 0.324 & \textbf{0.234} & \textbf{0.320}  & 0.246 & 0.329&  0.256 &  0.339 & 0.417 & 0.420   \\
                                                        & 48 & 0.321 & 0.380 & \textbf{0.317} & \textbf{0.378}  & 0.331 & 0.386&  0.339 & 0.396  & 0.526  &  0.484  \\
                                                        & 96 &0.370 & 0.414 & \textbf{0.364} & \textbf{0.412}  & 0.378 & 0.419& 0.376  &  0.422 & 0.573  &  0.516  \\
                                                        & 288 & 0.454 & \textbf{0.474} & \textbf{0.451} & \textbf{0.474} & 0.472 & 0.486 &  0.464 & 0.484  &  0.648 &  0.577  \\
                                                        & 672 & 0.610 & 0.569 & \textbf{0.591} & \textbf{0.560}  & 0.620 & 0.574& 0.608  &  0.566 &0.758   &   0.649 \\ \hline

        \multirow{5}{*}{\textbf{ETTm1} $\mathbf{\rightarrow}$\textbf{ETTh2}} & 24 & \textbf{0.336} & \textbf{0.425} & 0.342 & 0.428  & 0.447 & 0.502& 0.337  &  0.433  & 0.560  & 0.565  \\
                                                        & 48 & \textbf{0.563} & \textbf{0.554} & 0.568 & 0.557  & 0.699 & 0.637& 0.572  &  0.576  &  0.847 & 0.711  \\
                                                        & 168 & 1.378 & \textbf{0.913} & \textbf{1.367} & \textbf{0.913}  & 1.549 & 0.982& 1.470  & 0.947 & 2.327  & 1.206    \\
                                                        & 336 & 1.606 & 0.990 & \textbf{1.583} & \textbf{0.986}  & 1.749 & 1.042& 1.685  &  1.027  &  2.665 &  1.324 \\
                                                        & 720 & 1.925 & 1.101 & \textbf{1.906} & 1.106  & 1.971 & \textbf{1.092}&  1.890 & 1.092   &  2.690 &  1.365 \\ \hline
        
        \multirow{5}{*}{\textbf{ETTm1} $\mathbf{\rightarrow}$\textbf{ETTm2}} & 24 & 0.123 & 0.247 & 0.123 & 0.248  & 0.122 & 0.244& \textbf{0.120}  &  \textbf{0.238}  & 0.172  &  0.293 \\
                                                        & 48 & 0.180 & 0.307 & 0.181 & 0.307  & 0.183 & 0.305& \textbf{0.175}  & \textbf{0.295}   & 0.263  &  0.377 \\
                                                        & 96 & 0.287 & 0.391 & 0.287 & 0.392  & 0.294 & 0.394& \textbf{0.274}  & \textbf{0.375}  & 0.397  & 0.470   \\
                                                        & 288 & \textbf{0.681} & \textbf{0.623} & 0.687 & 0.627 & 0.723 & 0.652 & 0.725  & 0.638   &  0.897 & 0.733  \\
                                                        & 672 & 1.437 & \textbf{0.928} & \textbf{1.431} & 0.939  & 1.899 & 1.073& 1.746  & 1.005  & 2.185  &  1.144  \\ \hline
        
    \end{tabular}
    \label{table:transfer learning results}
\end{table*}

\subsection{Parameter Sensitivity Analysis}
In this section, we analyze the impact of hyperparameters on the performance of MFF-FTNet, including the number of dilated convolution blocks in the backbone encoder, denoted as $L^ \prime$, and the dimension of the intermediate feature representation, denoted as $D^ \prime$.
\subsubsection{The influence of $L^ \prime$}
\begin{figure}
    \centering
    % 第一行的两张图片
    \begin{minipage}{0.49\columnwidth}
        \centering
        \includegraphics[width=\linewidth]{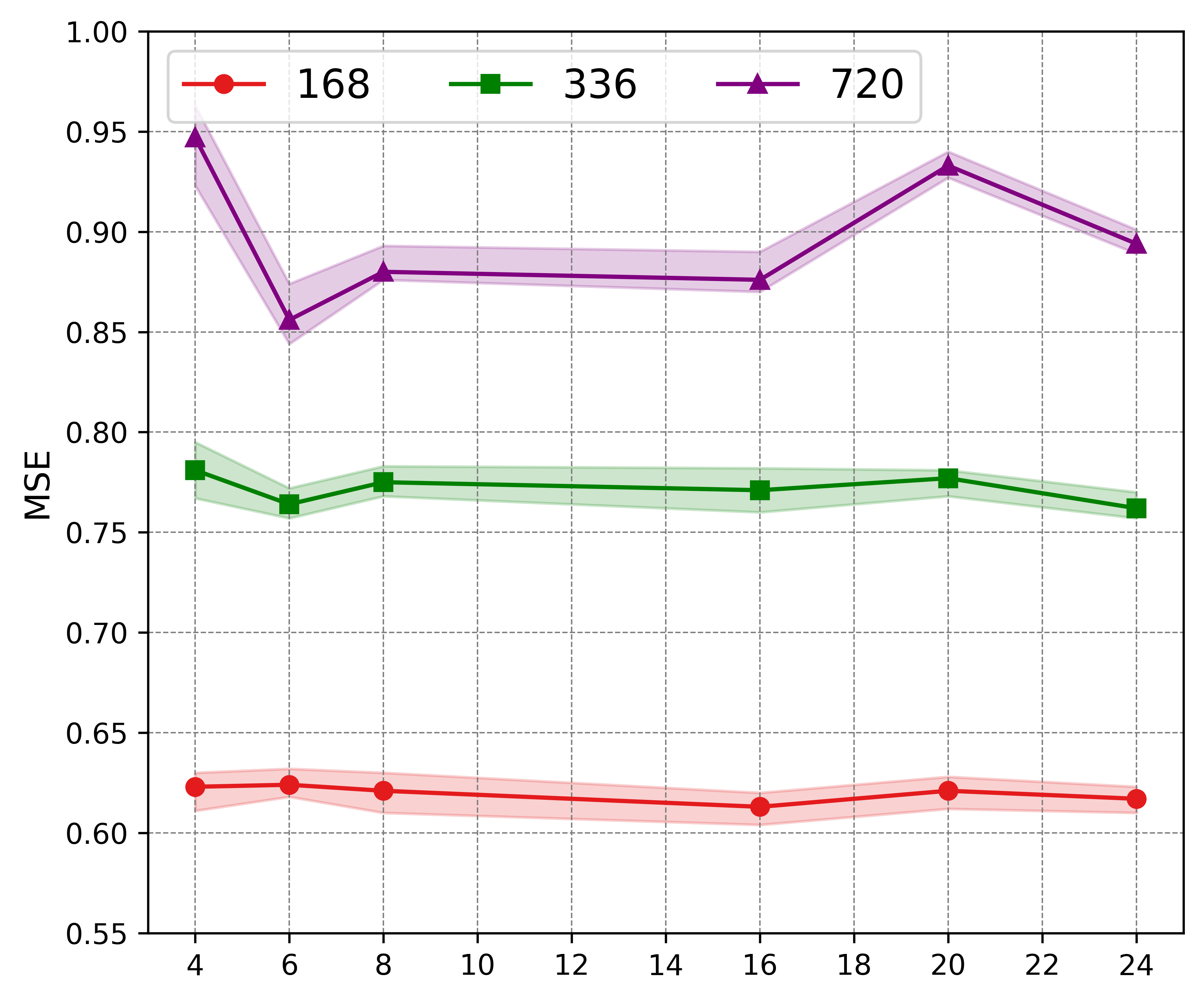}
        \caption*{ETTh1}
    \end{minipage}
    \hfill
    \begin{minipage}{0.49\columnwidth}
        \centering
        \includegraphics[width=\linewidth]{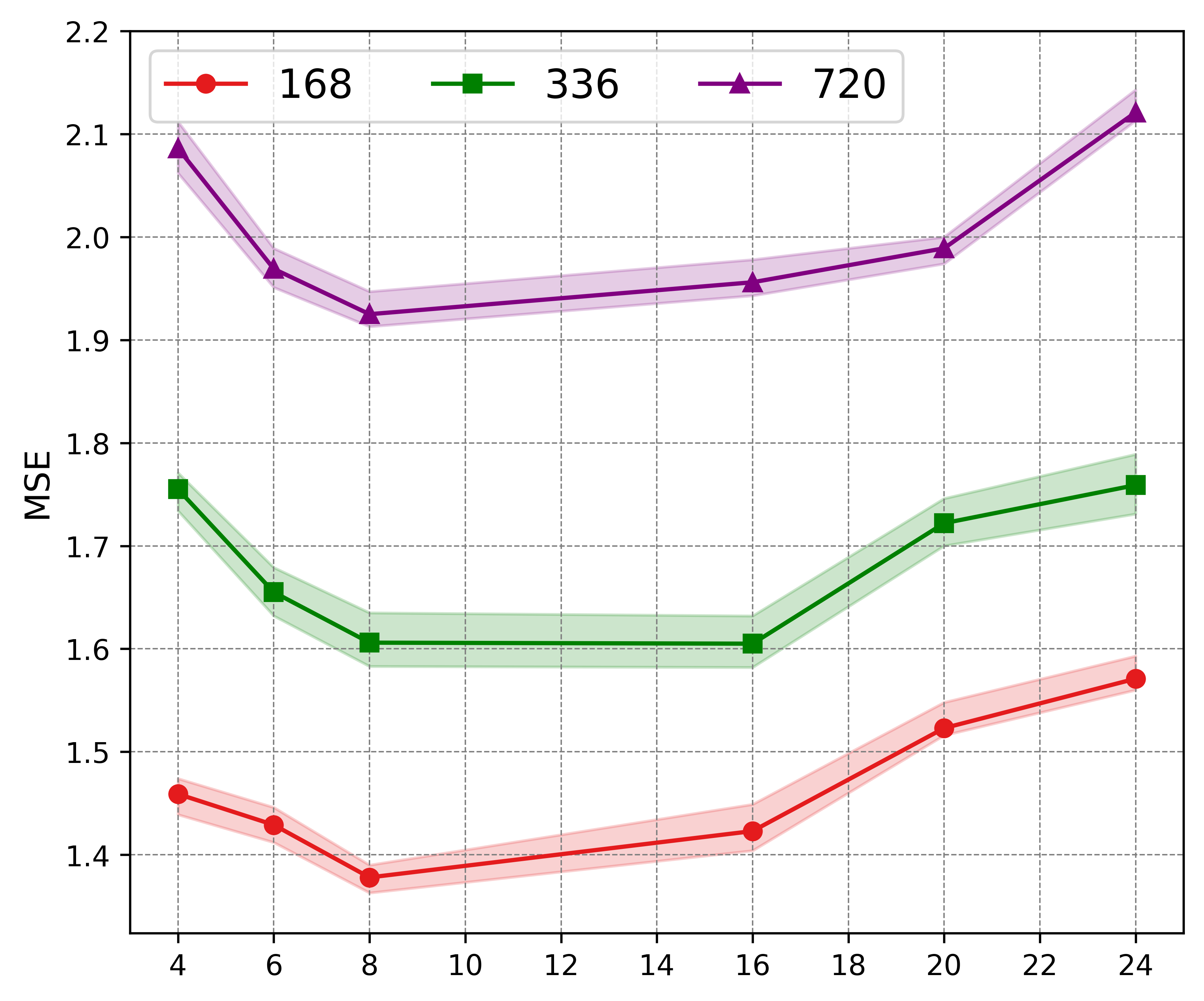}
        \caption*{ETTh2}
    \end{minipage}
    \vspace{0.1cm} % 控制上下图片间距
    % 第二行的两张图片
    \begin{minipage}{0.49\columnwidth}
        \centering
        \includegraphics[width=\linewidth]{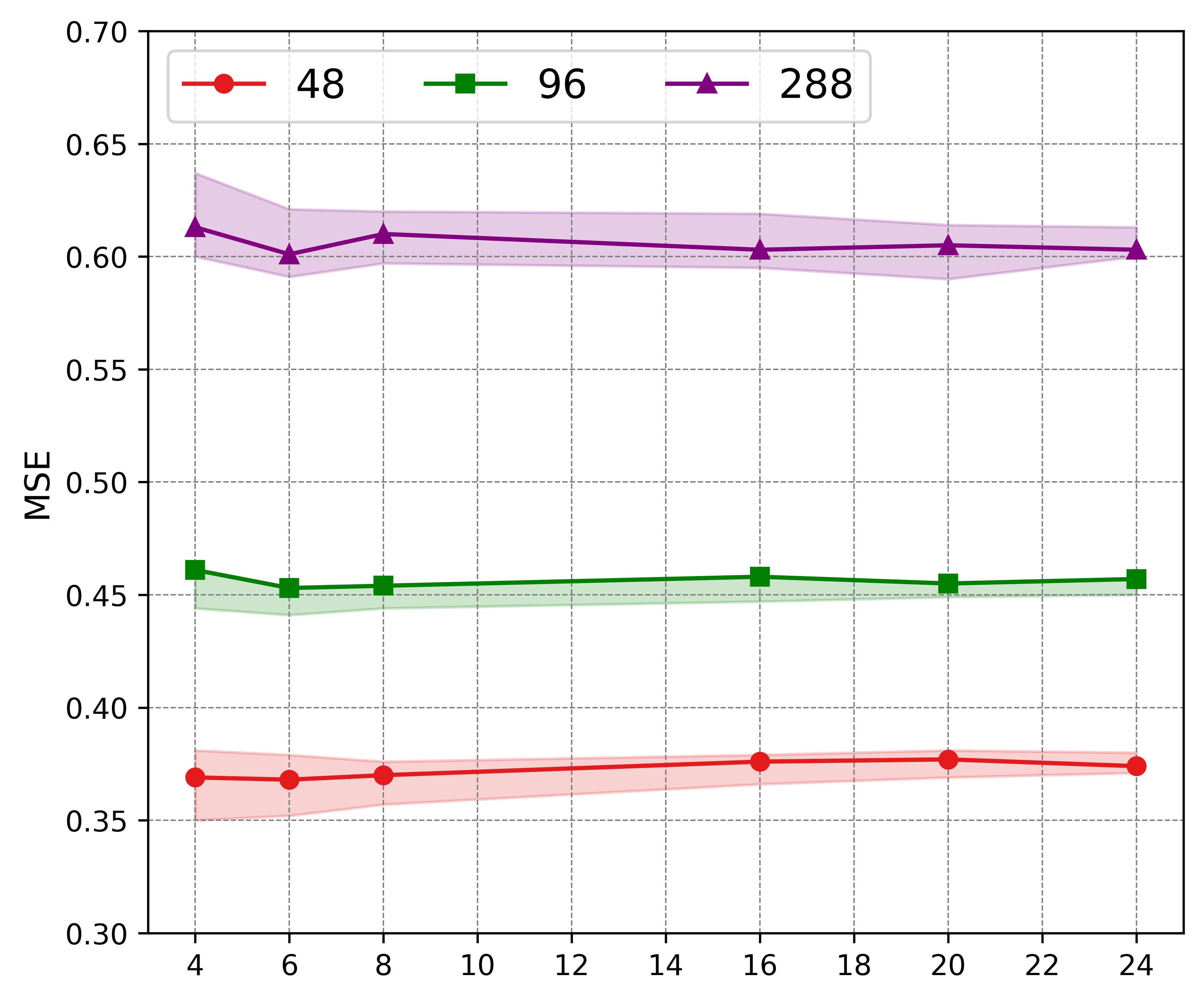}
        \caption*{ETTm1}
    \end{minipage}
    \hfill
    \begin{minipage}{0.49\columnwidth}
        \centering
        \includegraphics[width=\linewidth]{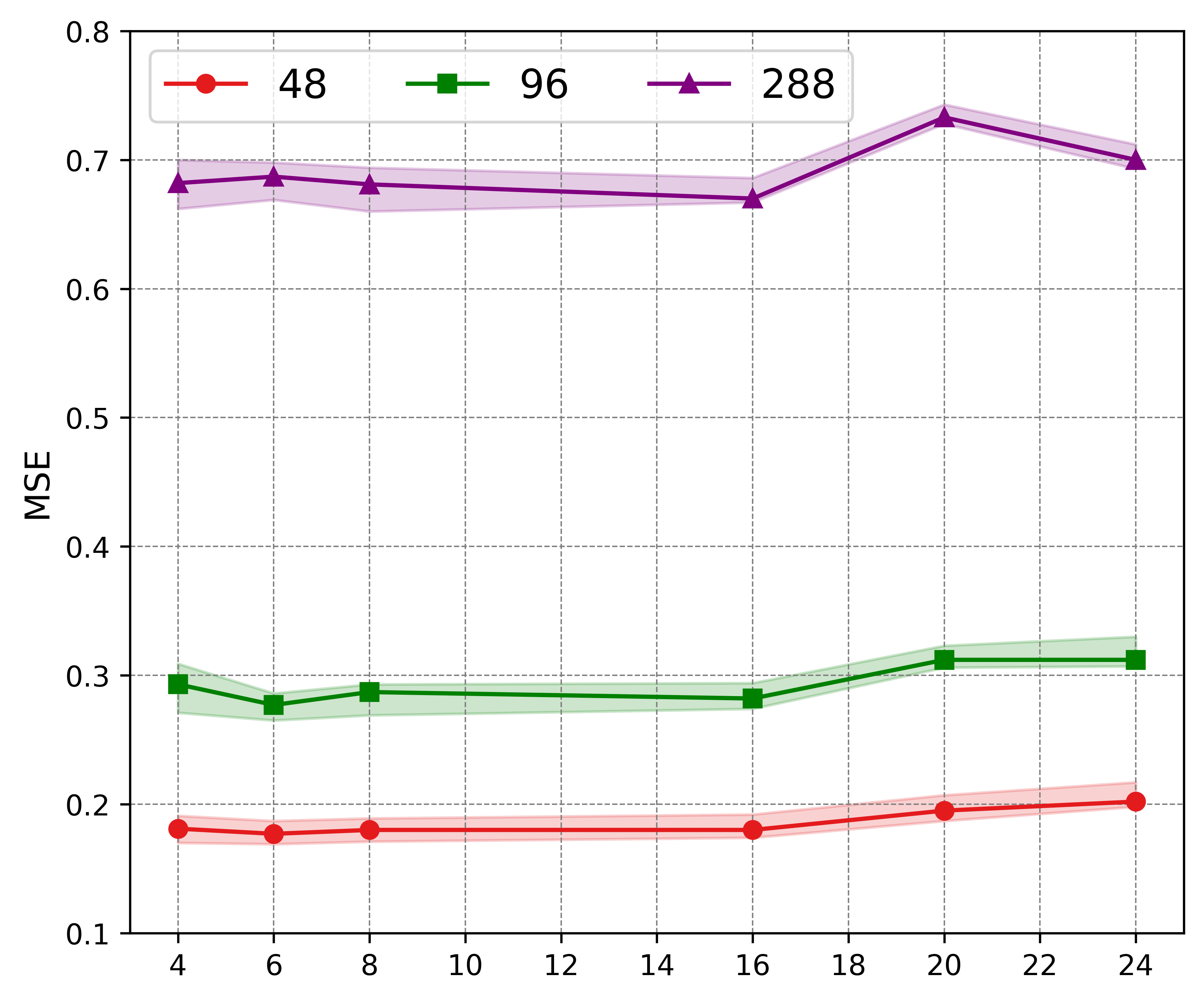}
        \caption*{ETTm2}
    \end{minipage}
    \caption{The impact of $L^ \prime$ on four different datasets for multivariate time series forecasting.}
    \label{fig:hyperparameter l analysis}
\end{figure}

This section analyzes how the number of dilated convolution blocks in the backbone affects prediction performance. Figure \ref{fig:hyperparameter l analysis} illustrates the model’s performance across different values of $L^ \prime \in \{4, 6, 8, 16, 20, 24\}$ on the ETTh1, ETTh2, ETTm1, and ETTm2 datasets for different forecasting horizons. As shown in the figure,  setting $L^ \prime=8$ achieves the best balance between performance and generalization across the datasets. With fewer blocks, such as $L^ \prime=4$ or $L^ \prime=6$, the model struggles to capture the complexity of both short- and long-term dependencies, leading to suboptimal performance. However, as $L^ \prime$ increases beyond 8, there is a diminishing return in accuracy, with deeper stacks (e.g., $L^ \prime=16$ or more) sometimes leading to overfitting. This overfitting can reduce the model’s ability to generalize effectively, as it may start focusing excessively on minor fluctuations in the training data rather than the underlying patterns necessary for robust forecasting.
\subsubsection{The influence of $D^ \prime$}
\begin{figure}
    \centering
    % 第一行的两张图片
    \begin{minipage}{0.49\columnwidth}
        \centering
        \includegraphics[width=\linewidth]{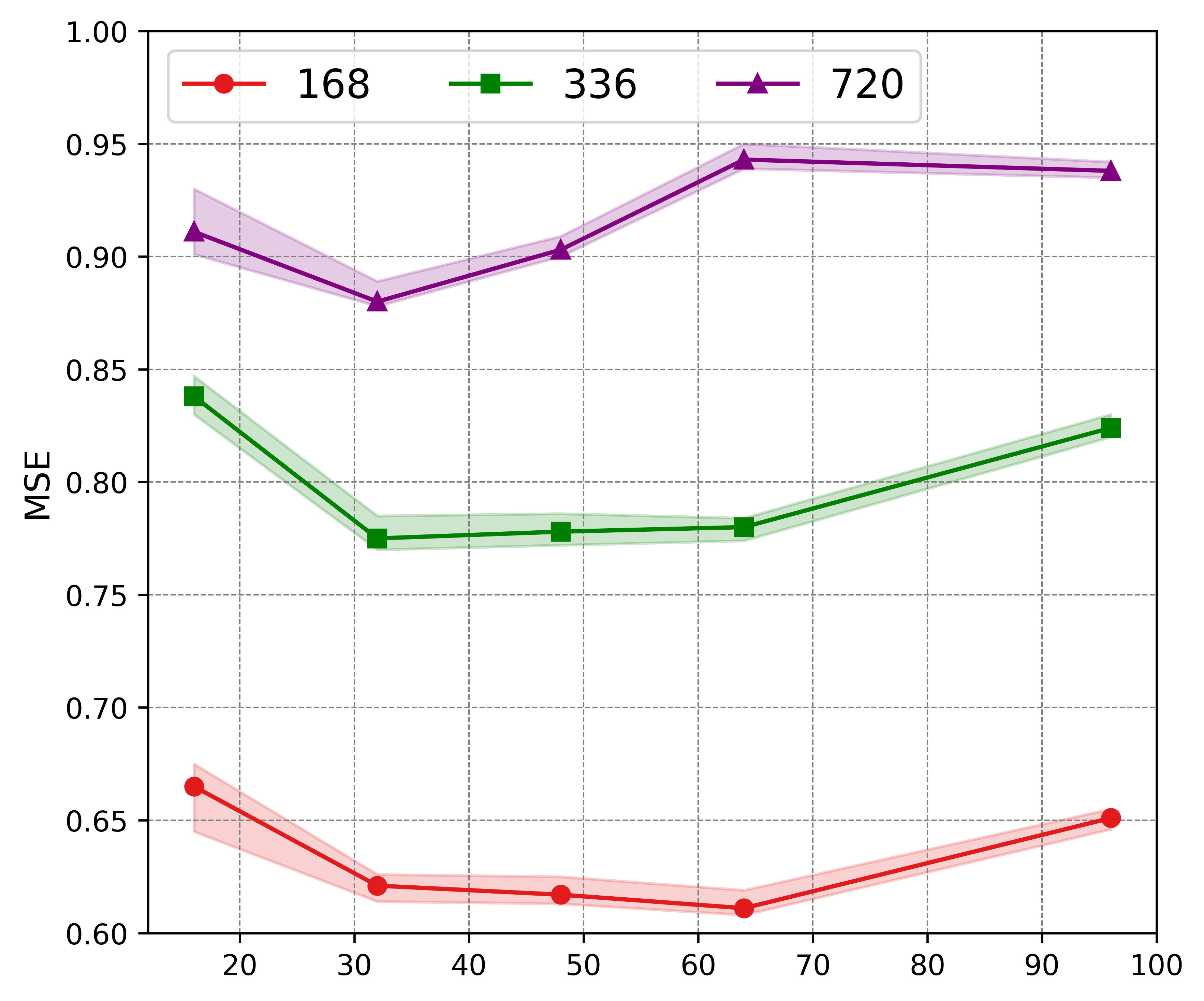}
        \caption*{ETTh1}
    \end{minipage}
    \hfill
    \begin{minipage}{0.49\columnwidth}
        \centering
        \includegraphics[width=\linewidth]{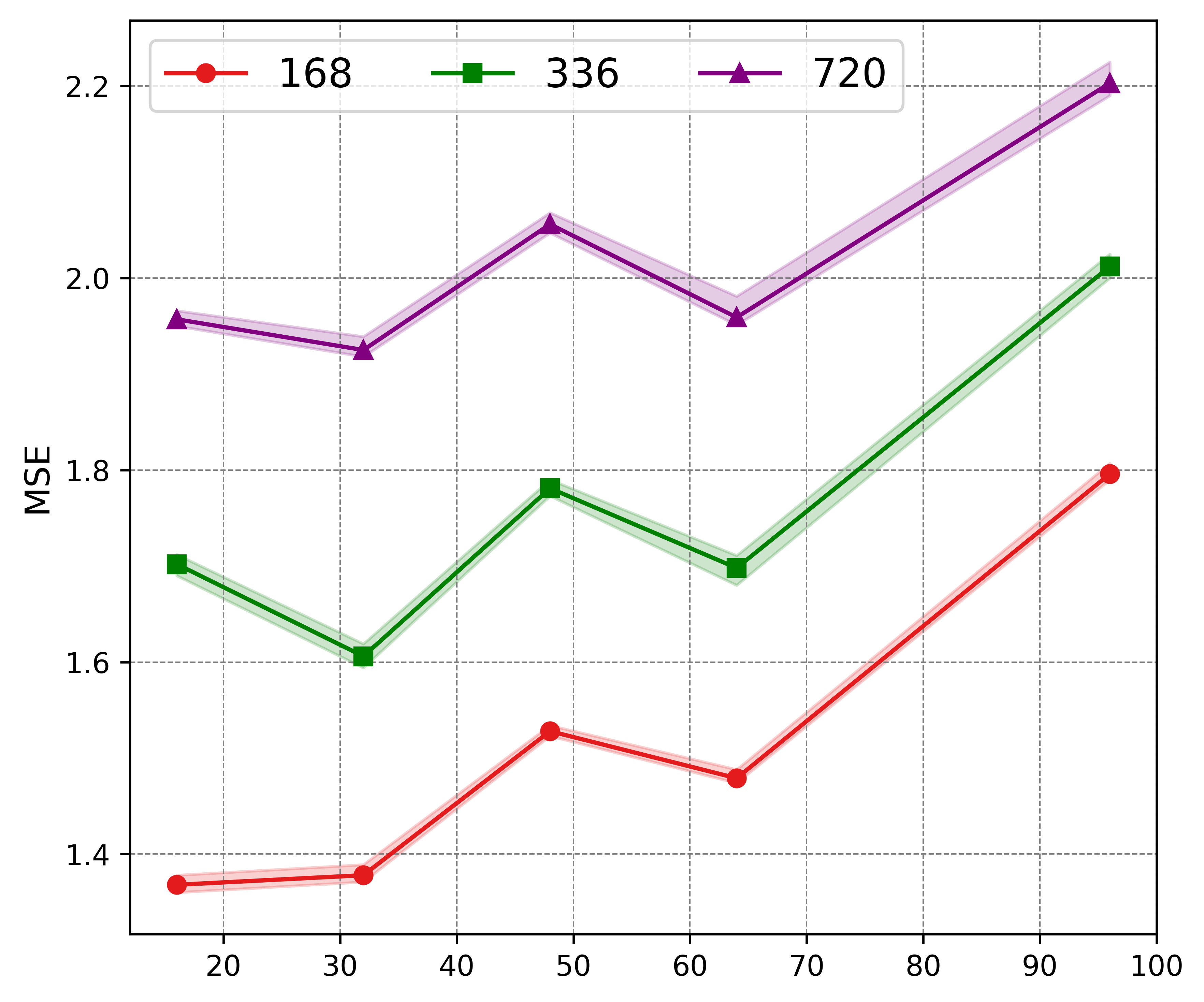}
        \caption*{ETTh2}
    \end{minipage}
    \vspace{0.1cm} % 控制上下图片间距
    % 第二行的两张图片
    \begin{minipage}{0.49\columnwidth}
        \centering
        \includegraphics[width=\linewidth]{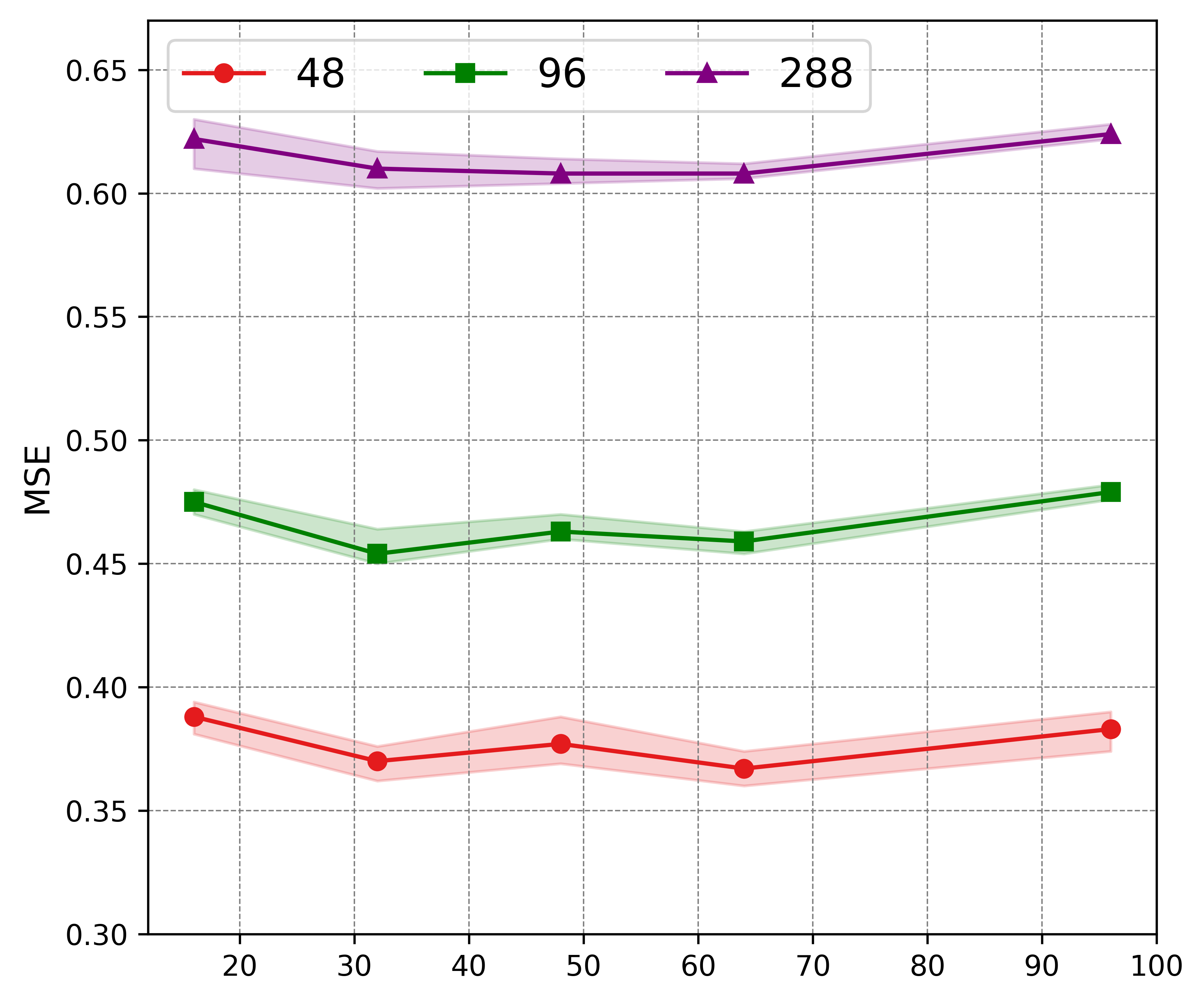}
        \caption*{ETTm1}
    \end{minipage}
    \hfill
    \begin{minipage}{0.49\columnwidth}
        \centering
        \includegraphics[width=\linewidth]{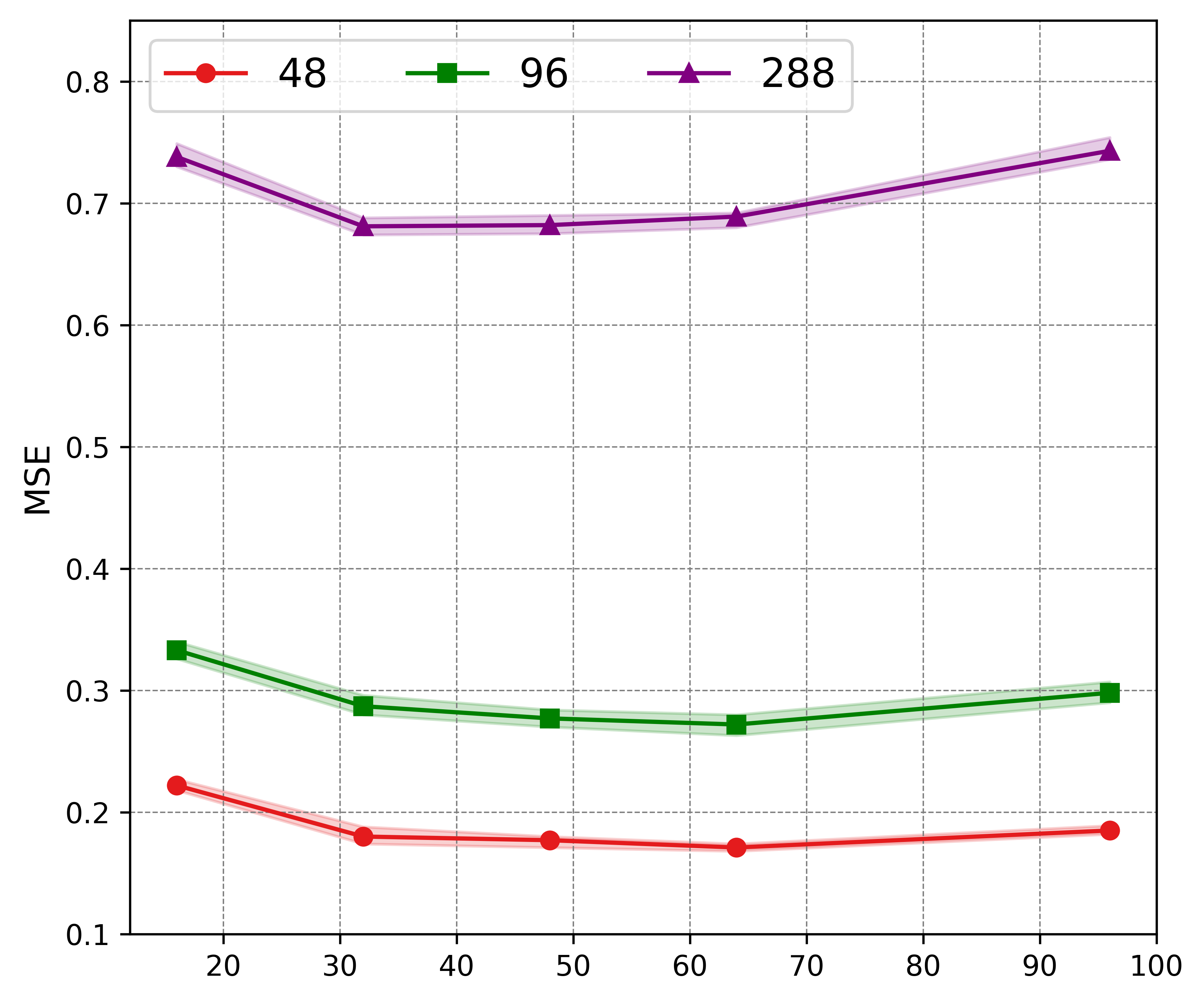}
        \caption*{ETTm2}
    \end{minipage}
    \caption{The impact of $D^ \prime$ on four different datasets for multivariate time series forecasting.}
    \label{fig:hyperparameter dim analysis}
\end{figure}

This section examines the effect of the hidden dimension, $D^ \prime$, in the backbone module on prediction performance. Figure \ref{fig:hyperparameter dim analysis} presents the model’s performance across varying hidden dimensions, $D^ \prime \in \{16, 32, 48, 64, 96\}$, across different horizons for the ETTh1, ETTh2, ETTm1, and ETTm2 datasets.
The results indicate that when $D^ \prime$ is set below 32, the model’s performance is generally weaker. Smaller hidden dimensions limit the model's capacity to capture complex temporal patterns, resulting in underfitting. This is particularly evident in longer prediction horizons, where capturing intricate relationships in the data is essential for accuracy. On the other hand, when $D^ \prime$ exceeds 32, there is a slight decline in performance, suggesting that overly large hidden dimensions lead to a more complex model that risks overfitting on the training data. This excessive complexity can hinder the model’s ability to generalize effectively to unseen data.

\begin{figure}
    \centering
    % 第一行的两张图片
    \begin{minipage}{0.49\columnwidth}
        \centering
        \includegraphics[width=\linewidth]{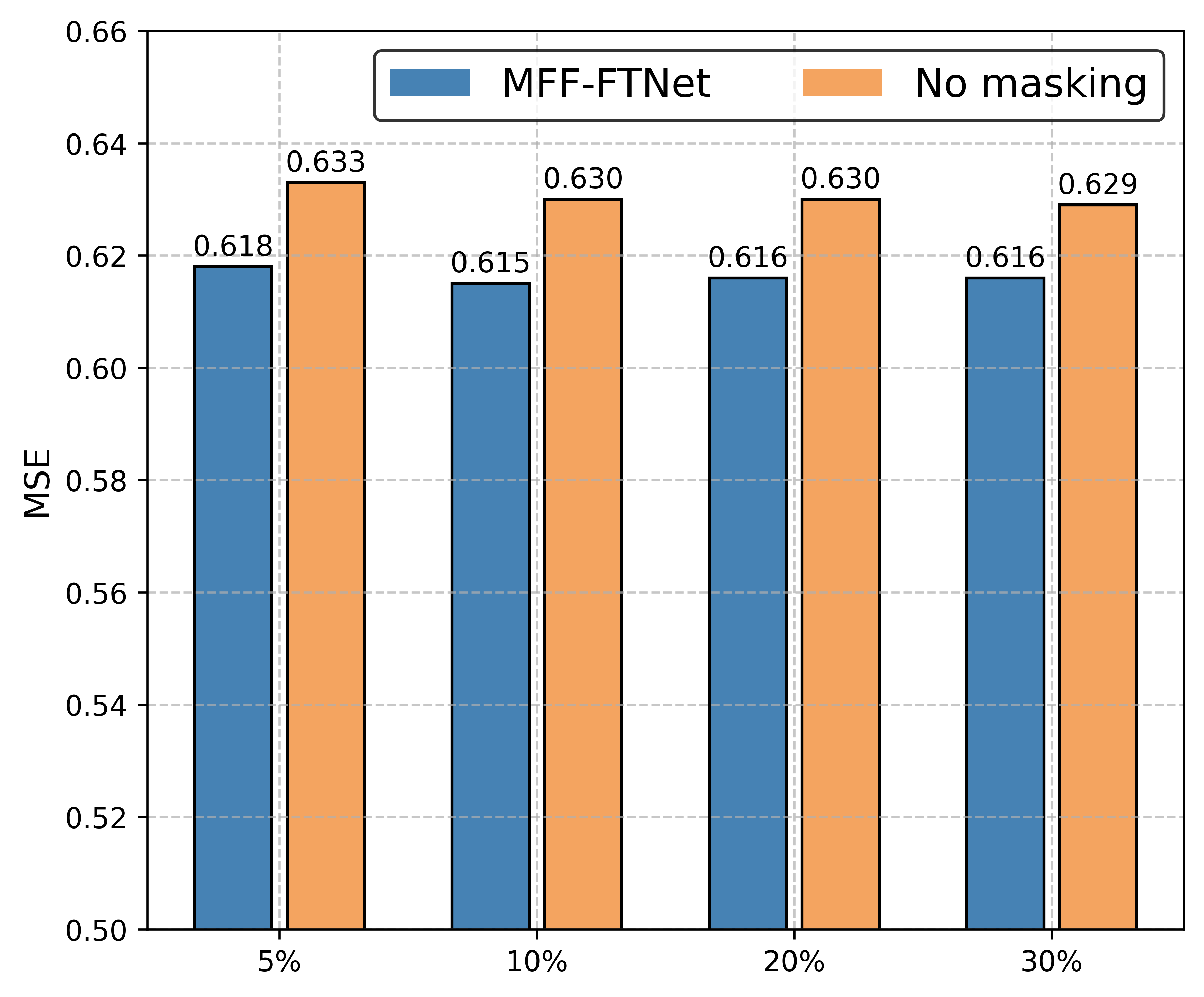}
        \caption*{ETTh1-168}
    \end{minipage}
    \hfill
    \begin{minipage}{0.49\columnwidth}
        \centering
        \includegraphics[width=\linewidth]{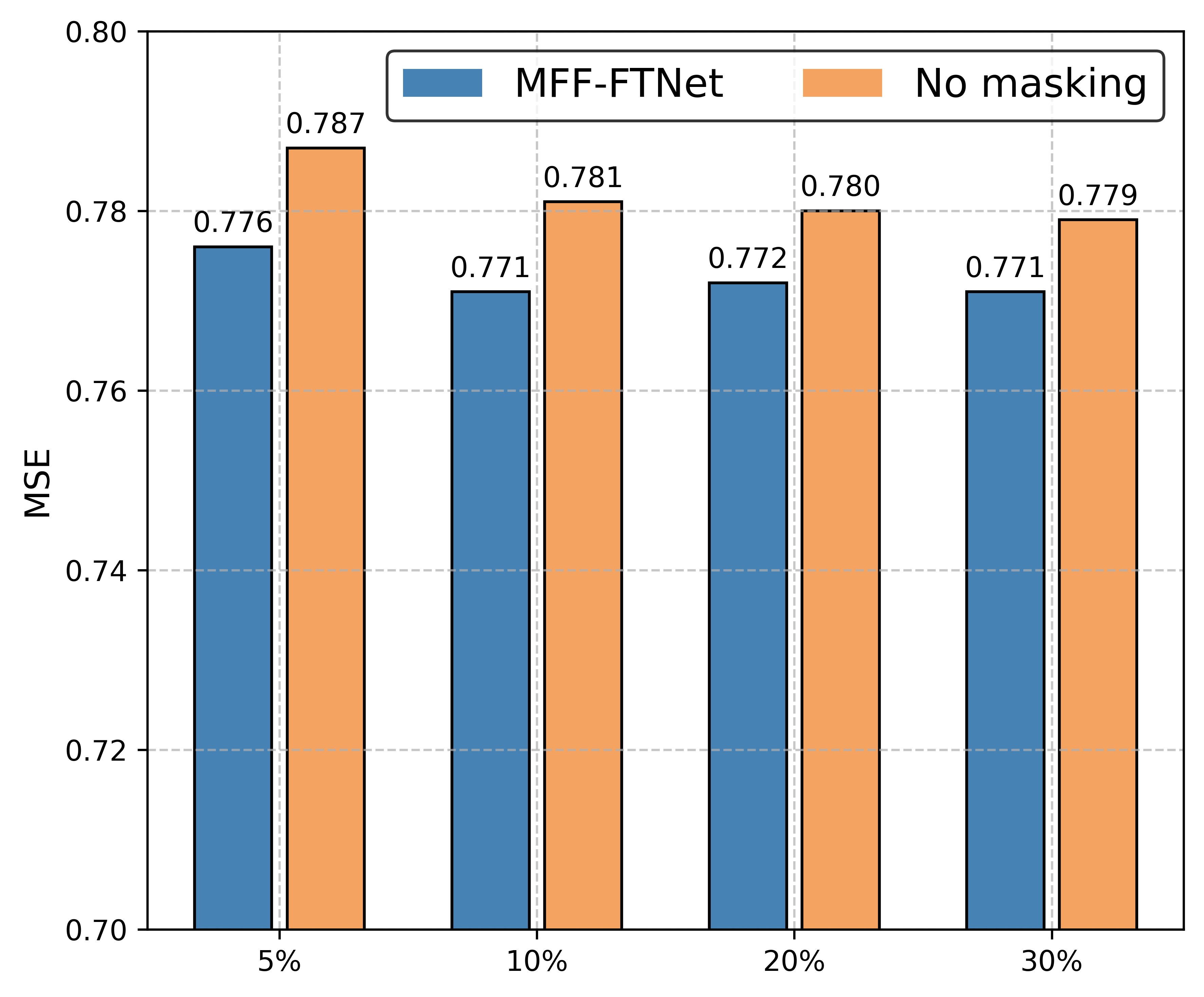}
        \caption*{ETTh1-336}
    \end{minipage}
    \vspace{0.1cm} % 控制上下图片间距
    % 第二行的两张图片
    \begin{minipage}{0.49\columnwidth}
        \centering
        \includegraphics[width=\linewidth]{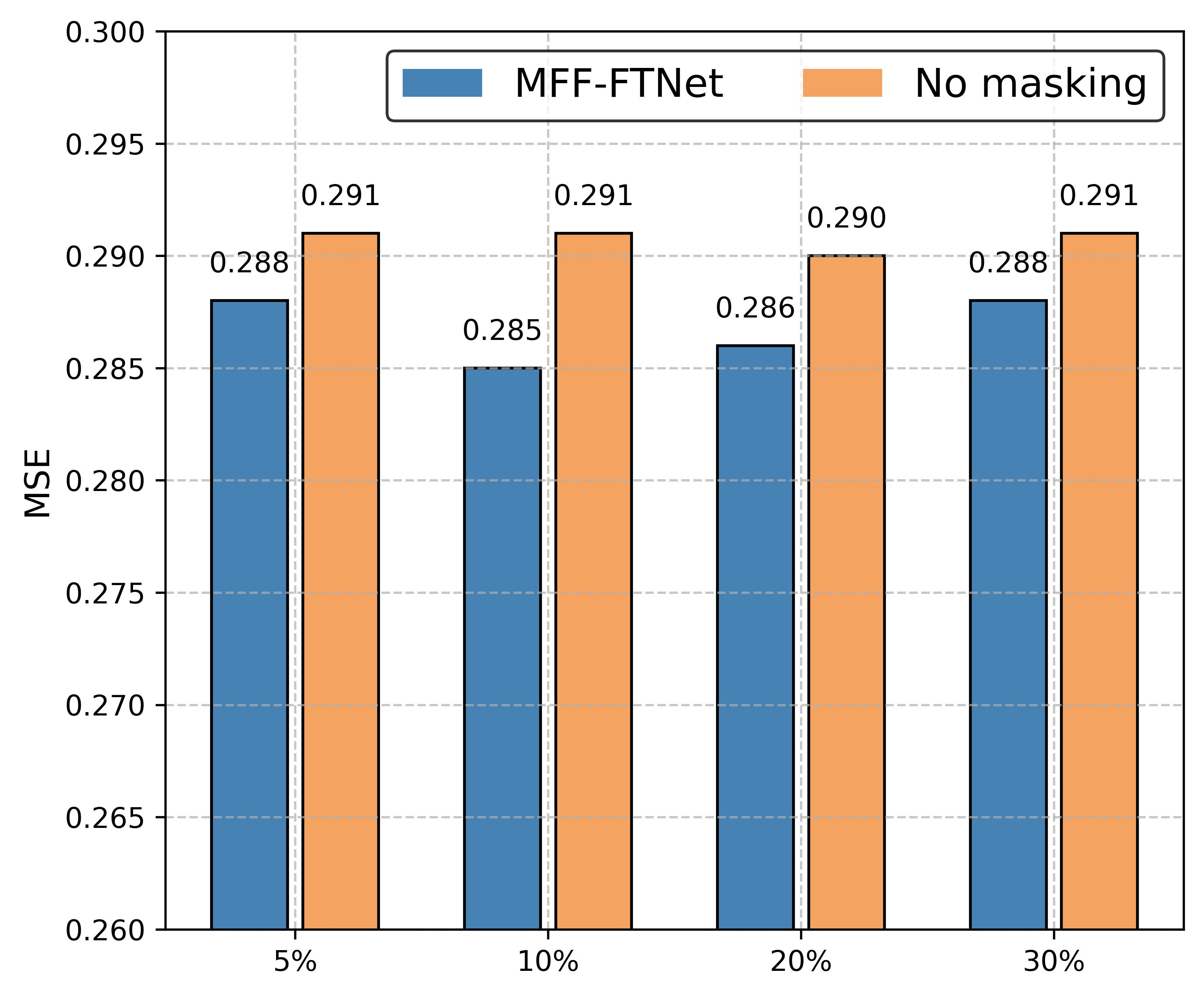}
        \caption*{ETTm2-96}
    \end{minipage}
    \hfill
    \begin{minipage}{0.49\columnwidth}
        \centering
        \includegraphics[width=\linewidth]{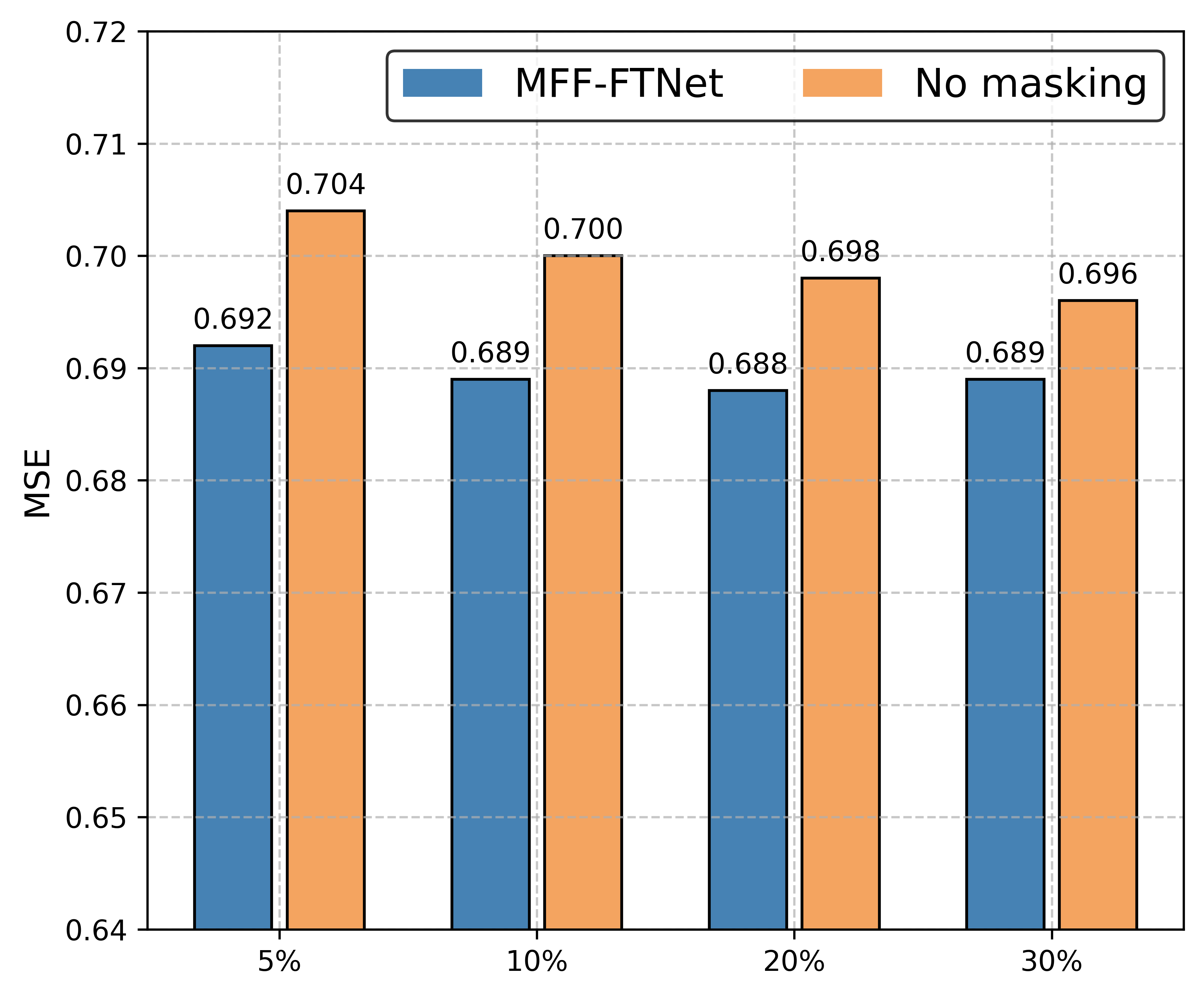}
        \caption*{ETTm2-288}
    \end{minipage}
    \caption{The study of robustness to noisy data on the ETTh1 and ETTm2 datasets.}
    \label{fig:robustness to noisy data}
\end{figure}
% ETTh1-336 heat map all dimension (14day)
% ETTm2-48 heat map all dimension (12hour)

\begin{figure}
    \centering
    \begin{minipage}{\columnwidth}
        \centering
        \includegraphics[width=\linewidth]{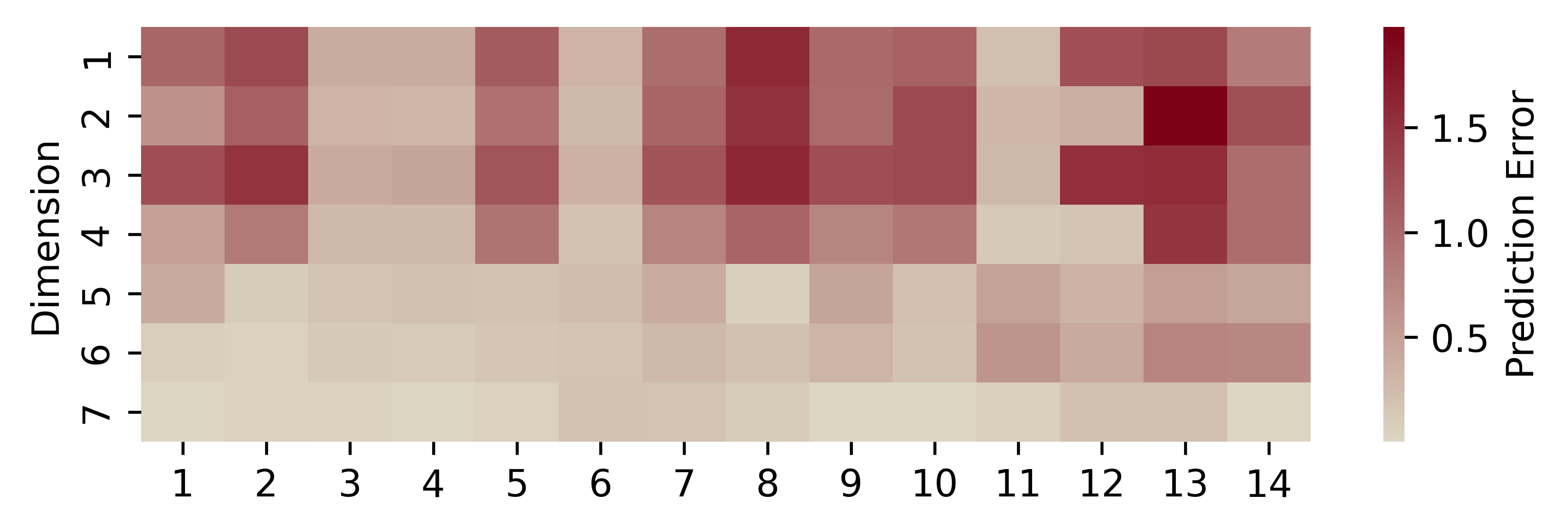}
        \caption*{Heat map for ETTh1 (14 days)}
    \end{minipage}
    \vspace{0.2cm} % 控制上下图片间距
    \begin{minipage}{\columnwidth}
        \centering
        \includegraphics[width=\linewidth]{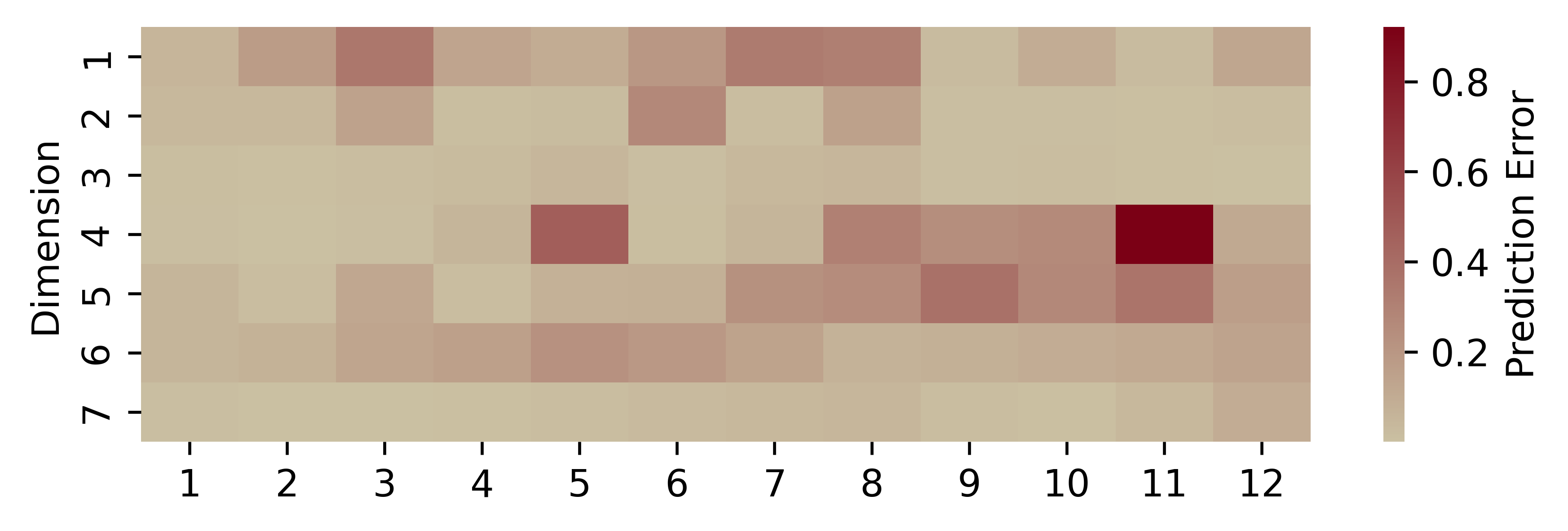}
        \caption*{Heat map for ETTm2 (12 hours)}
    \end{minipage}
    \caption{The heat map for the ETTh1 and ETTm2 datasets with different horizons.}
    \label{fig:heat map to noisy data}
\end{figure}

\setlength{\extrarowheight}{2pt}
\begin{table}
    \caption{The study of robustness to missing data on the ETTm1 and ETTm2 datasets.}
    \centering
    \begin{tabular}{cccccc}
        \hline
        \multicolumn{2}{l}{\textbf{Horizon}} & \multicolumn{2}{c}{\textbf{336}} & \multicolumn{2}{c}{\textbf{720}}\\ \hline
        \multicolumn{2}{l}{\textbf{Metric}} & \textbf{MSE} & \textbf{MAE} & \textbf{MSE} & \textbf{MAE} \\ \hline
        \multirow{4}{*}{\textbf{ETTh1}} & None & 0.775 & 0.664 & 0.880 & 0.737   \\
                                        & 10\%  & 0.779 & 0.666 & 0.887 & 0.740 \\
                                        & 20\%  & 0.781 & 0.667 & 0.890 & 0.741   \\
                                        & 30\%  & 0.783 & 0.668 & 0.893 & 0.743 \\
                                        & 40\%  & 0.783 & 0.668 & 0.895 & 0.743 \\ \hline
        \multicolumn{2}{l}{\textbf{pred length}} & \multicolumn{2}{c}{\textbf{96}} & \multicolumn{2}{c}{\textbf{288}}\\ \hline
        \multicolumn{2}{l}{\textbf{Metric}} & \textbf{MSE} & \textbf{MAE} & \textbf{MSE} & \textbf{MAE} \\ \hline
        \multirow{4}{*}{\textbf{ETTm2}} & None& 0.287 & 0.391 & 0.681 &  0.623 \\
                                        & 10\%  & 0.289 & 0.394 & 0.690 & 0.629 \\
                                        & 20\%  & 0.289 & 0.394 & 0.689 & 0.628   \\
                                        & 30\%  & 0.289 & 0.393 & 0.687 & 0.627 \\
                                        & 40\%  & 0.288 & 0.393 & 0.683 & 0.625 \\ \hline
        \end{tabular}
        \label{table:robustness to missing data}
\end{table}

\subsection{Transfer Learning}
Transfer learning aims to leverage the knowledge acquired during pretraining to enable rapid adaptation to new tasks across different domains, facilitating quicker generalization, and lowering training costs. We conducted two transfer learning experiments to evaluate MFF-FTNet's adaptability. The first experiment involved cross-domain transfer: pretraining on the WTH dataset and fine-tuning on the ETTh1 and ETTm1 datasets. The second experiment involved in-domain transfer: pretraining on the ETTm1 dataset and fine-tuning on the ETTh2 and ETTm2 datasets. Both experiments used the same parameters as in previous sections, with 600 epochs for pretraining and 300 epochs for fine-tuning. For comparison, we included CoST\cite{woo2022cost}, AutoTCL\cite{zheng2024parametric}, and T-Rep\cite{fraikin2023t} as baseline methods.

Table \ref{table:transfer learning results} illustrates that MFF-FTNet consistently outperforms baseline models, whether employing self-supervised training or a pretraining and fine-tuning approach. Notably, the pretraining and fine-tuning setup generally yields better prediction accuracy than self-supervised learning alone, with the most pronounced improvements observed on the ETTh1 and ETTm2 datasets. This suggests that MFF-FTNet effectively builds rich feature representations during pretraining, which facilitates quicker convergence and enhances its capacity to capture relevant task-specific features upon transfer.

\subsection{Robustness Study}
In real-world applications, data often contain noise or missing values due to measurement errors or environmental interference. To simulate these scenarios, we evaluated MFF-FTNet’s robustness by introducing Gaussian noise and varying levels of sparsity in the training data.

To examine robustness against noise, we added Gaussian noise to the ETTh1 and ETTm2 datasets with a mean and standard deviation of 10. Noise was added at proportions of 5\%, 10\%, 20\%, and 30\%. We also compared the results to a model variant without the frequency masking operation to test its effectiveness. Figure \ref{fig:robustness to noisy data} highlights two main observations. First, MFF-FTNet demonstrates strong resilience to noise, with prediction performance remaining stable across varying noise levels. In some cases, performance even improves slightly, particularly on the ETTh1 dataset, indicating that the model effectively handles noisy data. Second, the variant without masking operation shows significantly reduced performance in noisy conditions. This confirms that frequency masking operation effectively filters out noise unrelated to periodic patterns, enhancing prediction accuracy and robustness in noisy environments. To visually assess MFF-FTNet’s robustness under noisy conditions across time and feature dimensions, we generated heatmaps with a noise ratio of 40\% for the ETTh1 dataset (prediction length of 336, or 14 days) and the ETTm2 dataset (prediction length of 48, or 12 hours), as shown in Figure \ref{fig:heat map to noisy data}. 

To further assess robustness under data sparsity, we conducted experiments simulating missing values by randomly discarding different proportions of data points in the training set. For the ETTh1 and ETTm2 datasets, we tested sparsity levels by masking 10\%, 20\%, 30\%, and 40\% of the training data. Table \ref{table:robustness to missing data} shows that MFF-FTNet’s performance remains stable across all levels of sparsity, underscoring the model’s ability to maintain accurate predictions even with sparse data.

\subsection{Visualization Analysis}
\begin{figure} 
    \centering
    % 第一行的两张图片
    \begin{minipage}{0.49\columnwidth}
        \centering
        \includegraphics[width=\linewidth]{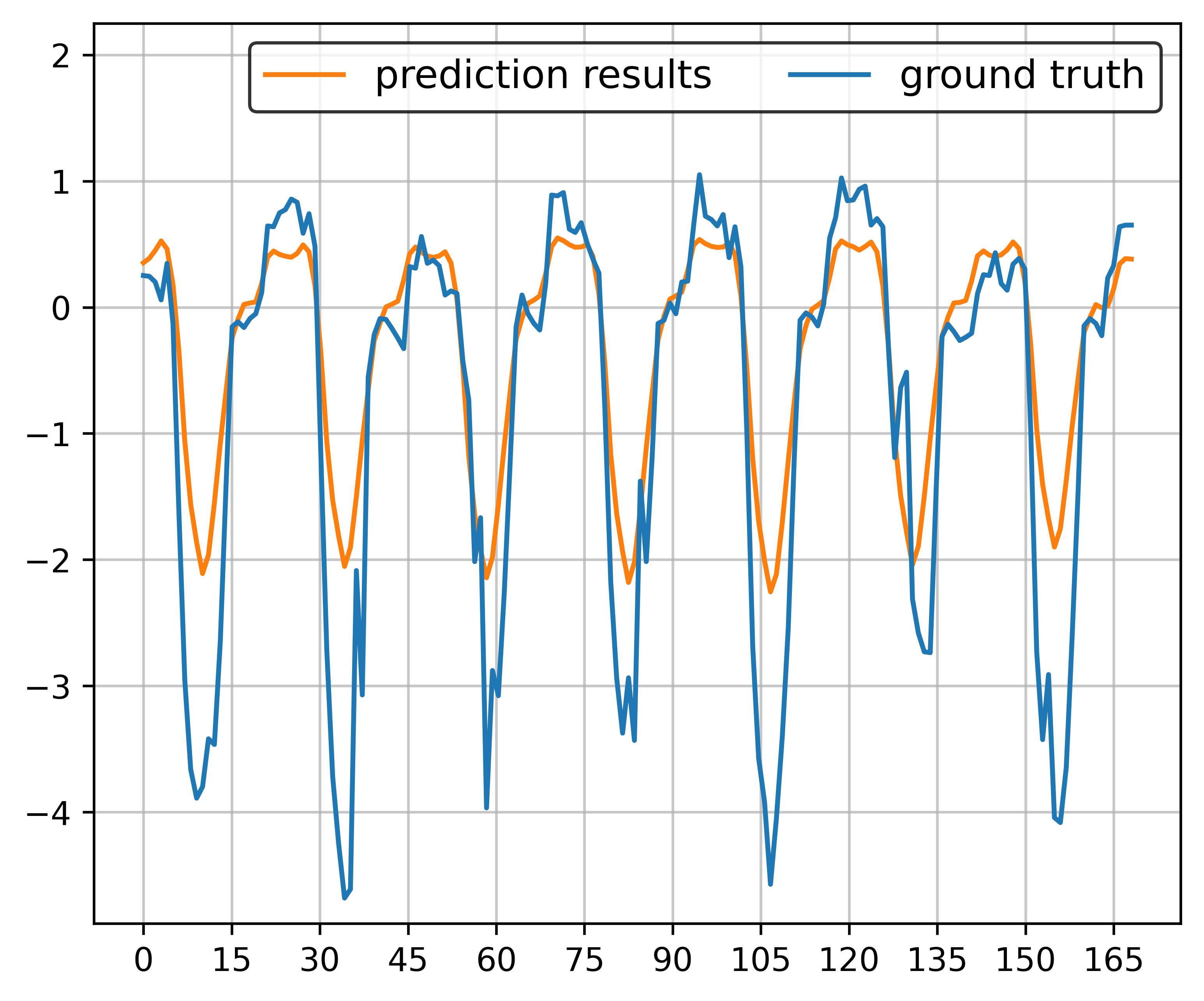}
        \caption*{ETTh1-168}
    \end{minipage}
    \hfill
    \begin{minipage}{0.49\columnwidth}
        \centering
        \includegraphics[width=\linewidth]{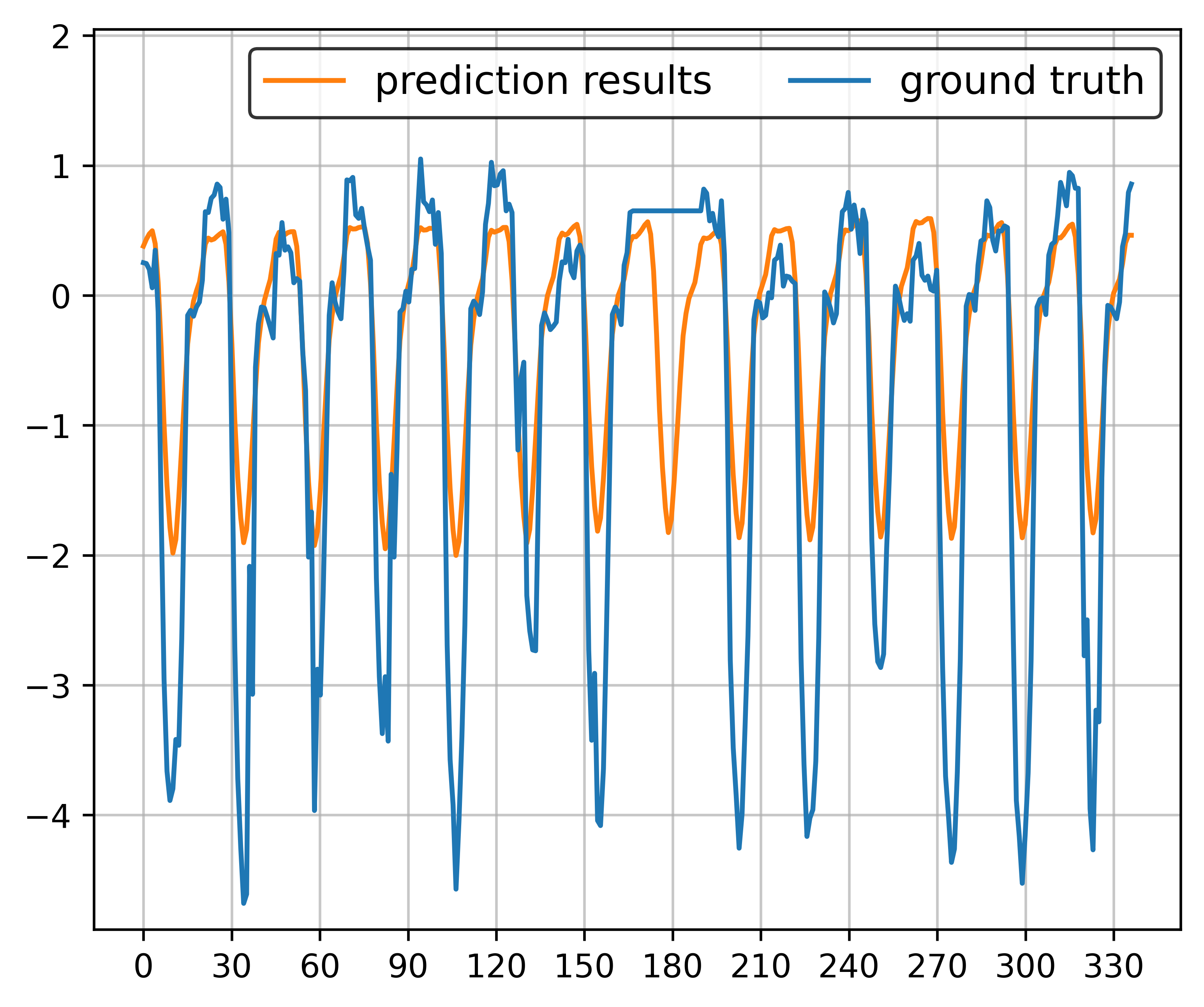}
        \caption*{ETTh1-336}
    \end{minipage}
    \vspace{0.1cm} % 控制上下图片间距
    % 第二行的两张图片
    \begin{minipage}{0.49\columnwidth}
        \centering
        \includegraphics[width=\linewidth]{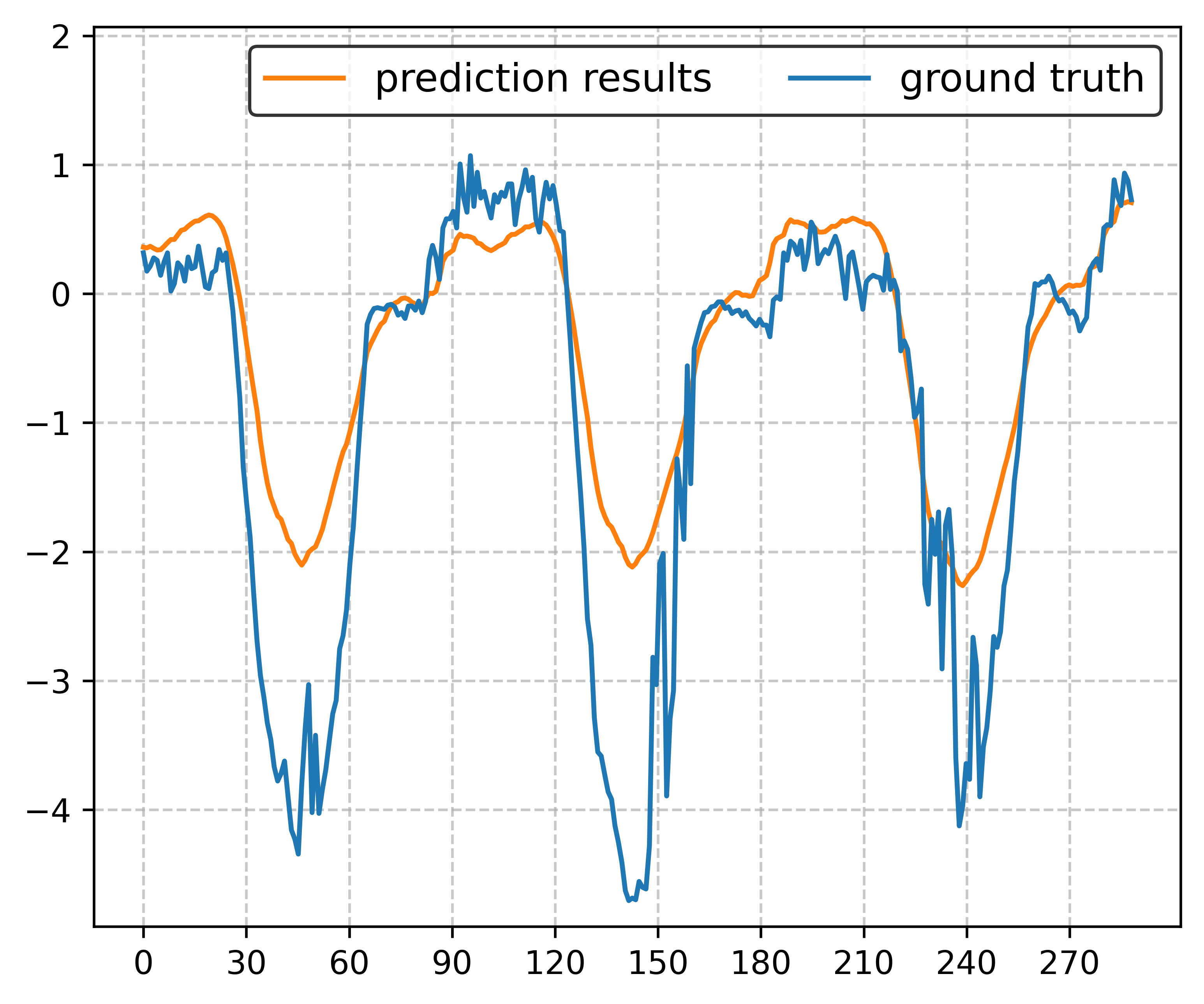}
        \caption*{ETTm1-288}
    \end{minipage}
    \hfill
    \begin{minipage}{0.49\columnwidth}
        \centering
        \includegraphics[width=\linewidth]{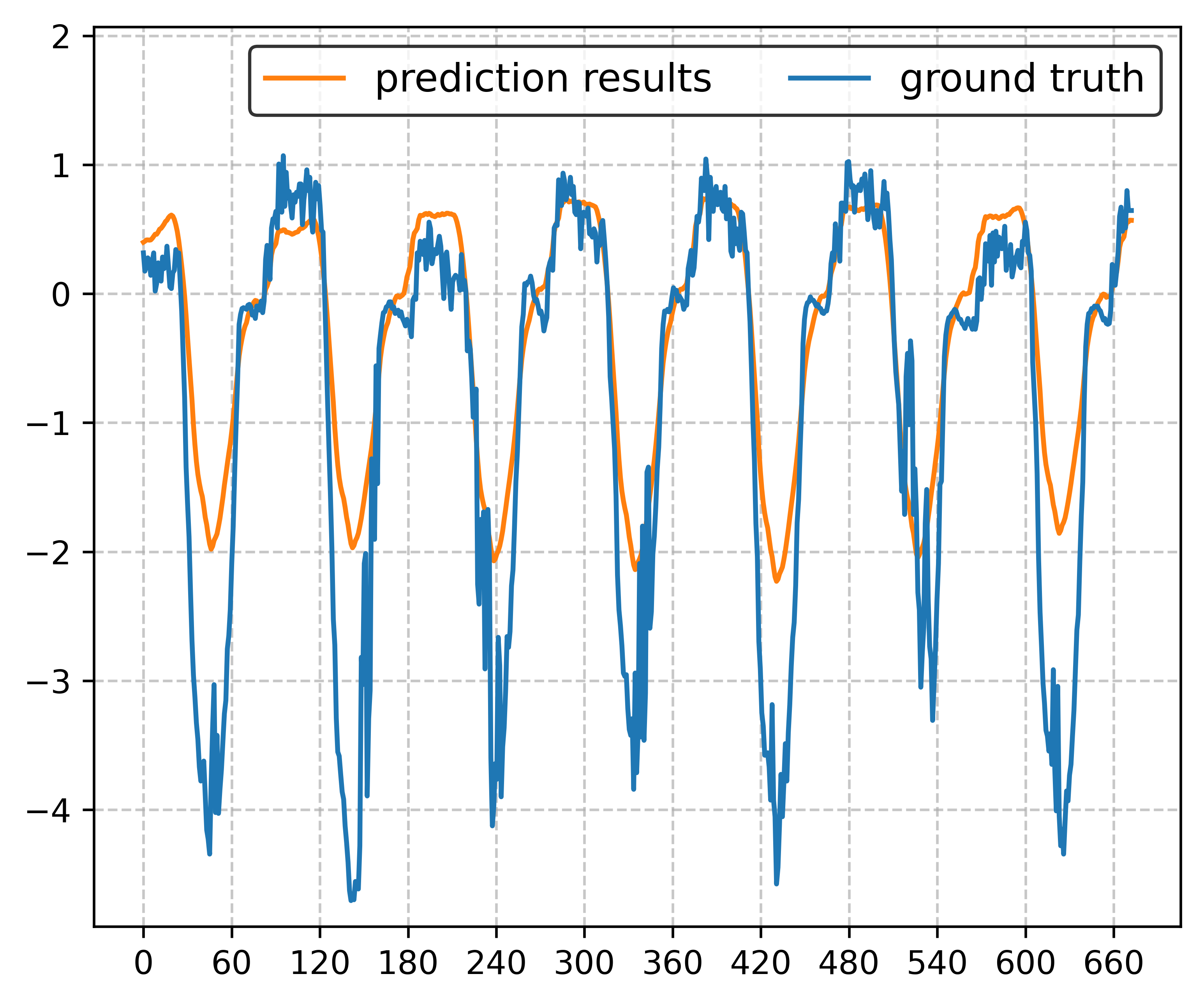}
        \caption*{ETTm1-672}
    \end{minipage}
    % \captionsetup[figure]{labelformat=default}
    \caption{Visualization results of MFF-FTNet for long-term prediction on ETTh1 and ETTm1 datasets.}
    \label{fig:Visualization}
\end{figure}

To illustrate the prediction accuracy of MFF-FTNet, we conducted a visualization analysis on the ETTh1 and ETTm1 datasets, comparing predicted values with actual outcomes. A sample of these predictions with different  horizons is shown in Figure \ref{fig:Visualization}.
For the ETTh1 dataset, the figure demonstrates that the predictions of MFF-FTNet align closely with the actual values, with only minor deviations. Peaks and troughs in the data are well-captured, indicating the model’s strong capability to recognize and follow periodic patterns within the series. On the ETTm1 dataset, MFF-FTNet similarly tracks the overarching trends and periodical behaviors, although some slight deviations are noted, particularly for predictions with a greater horizon, such as ETTm1-672. Despite these minor discrepancies, the model preserves the overall shape and direction of the time series data well.
\section{Conclusion}
In this paper, we address the challenges in time series forecasting, particularly in capturing both high-level periodic patterns and local temporal features while handling noise, multi-scale dependencies, and complex forecasting tasks. To tackle these issues, we propose MFF-FTNet, a novel framework that combines contrastive learning with multi-scale frequency and time-domain feature extraction. By integrating a Frequency-Aware Contrastive Module (FACM) and a Complementary Time Domain Contrastive Module (CTCM), MFF-FTNet enhances the model's ability to extract rich features from both frequency and time domains, making it more robust to noise and capable of detecting multi-scale patterns.
%We introduce adaptive frequency selection and dual-frequency contrastive loss in the FACM, improving the model's noise resilience and adaptability to non-stationary data. Additionally, the CTCM leverages multi-scale convolutions and feature fusion to capture intricate temporal dependencies, enhancing the overall forecasting accuracy. The combination of these two modules allows MFF-FTNet to model complex time series data effectively, providing a unified and enriched representation of both frequency and time-domain features.
Experimental results across five real-world datasets demonstrate that MFF-FTNet significantly outperforms state-of-the-art models, achieving substantial performance improvements. Future work could focus on further optimizing the computational efficiency of MFF-FTNet, particularly in handling larger datasets and real-time forecasting applications. 
\vspace{12pt}
\bibliographystyle{IEEEtran}
\bibliography{ref}

\begin{thebibliography}{10}
\providecommand{\url}[1]{#1}
\csname url@samestyle\endcsname
\providecommand{\newblock}{\relax}
\providecommand{\bibinfo}[2]{#2}
\providecommand{\BIBentrySTDinterwordspacing}{\spaceskip=0pt\relax}
\providecommand{\BIBentryALTinterwordstretchfactor}{4}
\providecommand{\BIBentryALTinterwordspacing}{\spaceskip=\fontdimen2\font plus
\BIBentryALTinterwordstretchfactor\fontdimen3\font minus \fontdimen4\font\relax}
\providecommand{\BIBforeignlanguage}[2]{{%
\expandafter\ifx\csname l@#1\endcsname\relax
\typeout{** WARNING: IEEEtran.bst: No hyphenation pattern has been}%
\typeout{** loaded for the language `#1'. Using the pattern for}%
\typeout{** the default language instead.}%
\else
\language=\csname l@#1\endcsname
\fi
#2}}
\providecommand{\BIBdecl}{\relax}
\BIBdecl

\bibitem{zhao2023doubleadapt}
L.~Zhao, S.~Kong, and Y.~Shen, ``Doubleadapt: A meta-learning approach to incremental learning for stock trend forecasting,'' in \emph{Proceedings of the 29th ACM SIGKDD Conference on Knowledge Discovery and Data Mining}, 2023, pp. 3492--3503.

\bibitem{cao2022ai}
L.~Cao, ``Ai in finance: challenges, techniques, and opportunities,'' \emph{ACM Computing Surveys (CSUR)}, vol.~55, no.~3, pp. 1--38, 2022.

\bibitem{houssein2022efficient}
E.~H. Houssein, M.~Dirar, L.~Abualigah, and W.~M. Mohamed, ``An efficient equilibrium optimizer with support vector regression for stock market prediction,'' \emph{Neural computing and applications}, pp. 1--36, 2022.

\bibitem{bi2023accurate}
K.~Bi, L.~Xie, H.~Zhang, X.~Chen, X.~Gu, and Q.~Tian, ``Accurate medium-range global weather forecasting with 3d neural networks,'' \emph{Nature}, vol. 619, no. 7970, pp. 533--538, 2023.

\bibitem{zhuang2023data}
D.~Zhuang, V.~J. Gan, Z.~D. Tekler, A.~Chong, S.~Tian, and X.~Shi, ``Data-driven predictive control for smart hvac system in iot-integrated buildings with time-series forecasting and reinforcement learning,'' \emph{Applied Energy}, vol. 338, p. 120936, 2023.

\bibitem{jiang2023spatio}
R.~Jiang, Z.~Wang, J.~Yong, P.~Jeph, Q.~Chen, Y.~Kobayashi, X.~Song, S.~Fukushima, and T.~Suzumura, ``Spatio-temporal meta-graph learning for traffic forecasting,'' in \emph{Proceedings of the AAAI conference on artificial intelligence}, vol.~37, no.~7, 2023, pp. 8078--8086.

\bibitem{morid2023time}
M.~A. Morid, O.~R.~L. Sheng, and J.~Dunbar, ``Time series prediction using deep learning methods in healthcare,'' \emph{ACM Transactions on Management Information Systems}, vol.~14, no.~1, pp. 1--29, 2023.

\bibitem{dudek2023std}
G.~Dudek, ``Std: a seasonal-trend-dispersion decomposition of time series,'' \emph{IEEE Transactions on Knowledge and Data Engineering}, vol.~35, no.~10, pp. 10\,339--10\,350, 2023.

\bibitem{tokgoz2018rnn}
A.~Tokg{\"o}z and G.~{\"U}nal, ``A rnn based time series approach for forecasting turkish electricity load,'' in \emph{2018 26th Signal processing and communications applications conference (SIU)}.\hskip 1em plus 0.5em minus 0.4em\relax IEEE, 2018, pp. 1--4.

\bibitem{wang2022ngcu}
J.~Wang, X.~Li, J.~Li, Q.~Sun, and H.~Wang, ``Ngcu: A new rnn model for time-series data prediction,'' \emph{Big Data Research}, vol.~27, p. 100296, 2022.

\bibitem{siami2019performance}
S.~Siami-Namini, N.~Tavakoli, and A.~S. Namin, ``The performance of lstm and bilstm in forecasting time series,'' in \emph{2019 IEEE International conference on big data (Big Data)}.\hskip 1em plus 0.5em minus 0.4em\relax IEEE, 2019, pp. 3285--3292.

\bibitem{yin2023u}
L.~Yin, L.~Wang, T.~Li, S.~Lu, J.~Tian, Z.~Yin, X.~Li, and W.~Zheng, ``U-net-lstm: time series-enhanced lake boundary prediction model,'' \emph{Land}, vol.~12, no.~10, p. 1859, 2023.

\bibitem{livieris2020cnn}
I.~E. Livieris, E.~Pintelas, and P.~Pintelas, ``A cnn--lstm model for gold price time-series forecasting,'' \emph{Neural computing and applications}, vol.~32, pp. 17\,351--17\,360, 2020.

\bibitem{jin2020prediction}
X.~Jin, X.~Yu, X.~Wang, Y.~Bai, T.~Su, and J.~Kong, ``Prediction for time series with cnn and lstm,'' in \emph{Proceedings of the 11th international conference on modelling, identification and control (ICMIC2019)}.\hskip 1em plus 0.5em minus 0.4em\relax Springer, 2020, pp. 631--641.

\bibitem{yue2022ts2vec}
Z.~Yue, Y.~Wang, J.~Duan, T.~Yang, C.~Huang, Y.~Tong, and B.~Xu, ``Ts2vec: Towards universal representation of time series,'' in \emph{Proceedings of the AAAI Conference on Artificial Intelligence}, vol.~36, no.~8, 2022, pp. 8980--8987.

\bibitem{fraikin2023t}
A.~Fraikin, A.~Bennetot, and S.~Allassonni{\`e}re, ``T-rep: Representation learning for time series using time-embeddings,'' \emph{arXiv preprint arXiv:2310.04486}, 2023.

\bibitem{zheng2023simts}
X.~Zheng, X.~Chen, M.~Sch{\"u}rch, A.~Mollaysa, A.~Allam, and M.~Krauthammer, ``Simts: rethinking contrastive representation learning for time series forecasting,'' \emph{arXiv preprint arXiv:2303.18205}, 2023.

\bibitem{zhou2021informer}
H.~Zhou, S.~Zhang, J.~Peng, S.~Zhang, J.~Li, H.~Xiong, and W.~Zhang, ``Informer: Beyond efficient transformer for long sequence time-series forecasting,'' in \emph{Proceedings of the AAAI conference on artificial intelligence}, vol.~35, no.~12, 2021, pp. 11\,106--11\,115.

\bibitem{nie2022time}
Y.~Nie, N.~H. Nguyen, P.~Sinthong, and J.~Kalagnanam, ``A time series is worth 64 words: Long-term forecasting with transformers,'' \emph{arXiv preprint arXiv:2211.14730}, 2022.

\bibitem{darban2025carla}
Z.~Z. Darban, G.~I. Webb, S.~Pan, C.~C. Aggarwal, and M.~Salehi, ``Carla: Self-supervised contrastive representation learning for time series anomaly detection,'' \emph{Pattern Recognition}, vol. 157, p. 110874, 2025.

\bibitem{woo2022cost}
G.~Woo, C.~Liu, D.~Sahoo, A.~Kumar, and S.~Hoi, ``Cost: Contrastive learning of disentangled seasonal-trend representations for time series forecasting,'' \emph{arXiv preprint arXiv:2202.01575}, 2022.

\bibitem{yi2024fouriergnn}
K.~Yi, Q.~Zhang, W.~Fan, H.~He, L.~Hu, P.~Wang, N.~An, L.~Cao, and Z.~Niu, ``Fouriergnn: Rethinking multivariate time series forecasting from a pure graph perspective,'' \emph{Advances in Neural Information Processing Systems}, vol.~36, 2024.

\bibitem{huang2023crossgnn}
Q.~Huang, L.~Shen, R.~Zhang, S.~Ding, B.~Wang, Z.~Zhou, and Y.~Wang, ``Crossgnn: Confronting noisy multivariate time series via cross interaction refinement,'' \emph{Advances in Neural Information Processing Systems}, vol.~36, pp. 46\,885--46\,902, 2023.

\bibitem{zhang2023crossformer}
Y.~Zhang and J.~Yan, ``Crossformer: Transformer utilizing cross-dimension dependency for multivariate time series forecasting,'' in \emph{The eleventh international conference on learning representations}, 2023.

\bibitem{chen2020simple}
T.~Chen, S.~Kornblith, M.~Norouzi, and G.~Hinton, ``A simple framework for contrastive learning of visual representations,'' in \emph{International conference on machine learning}.\hskip 1em plus 0.5em minus 0.4em\relax PMLR, 2020, pp. 1597--1607.

\bibitem{xie2020unsupervised}
Q.~Xie, Z.~Dai, E.~Hovy, T.~Luong, and Q.~Le, ``Unsupervised data augmentation for consistency training,'' \emph{Advances in neural information processing systems}, vol.~33, pp. 6256--6268, 2020.

\bibitem{you2020graph}
Y.~You, T.~Chen, Y.~Sui, T.~Chen, Z.~Wang, and Y.~Shen, ``Graph contrastive learning with augmentations,'' \emph{Advances in neural information processing systems}, vol.~33, pp. 5812--5823, 2020.

\bibitem{he2020momentum}
K.~He, H.~Fan, Y.~Wu, S.~Xie, and R.~Girshick, ``Momentum contrast for unsupervised visual representation learning,'' in \emph{Proceedings of the IEEE/CVF conference on computer vision and pattern recognition}, 2020, pp. 9729--9738.

\bibitem{eldele2021time}
E.~Eldele, M.~Ragab, Z.~Chen, M.~Wu, C.~K. Kwoh, X.~Li, and C.~Guan, ``Time-series representation learning via temporal and contextual contrasting,'' \emph{arXiv preprint arXiv:2106.14112}, 2021.

\bibitem{liu2024timesurl}
J.~Liu and S.~Chen, ``Timesurl: Self-supervised contrastive learning for universal time series representation learning,'' in \emph{Proceedings of the AAAI Conference on Artificial Intelligence}, vol.~38, no.~12, 2024, pp. 13\,918--13\,926.

\bibitem{WuHLZ0L23}
H.~Wu, T.~Hu, Y.~Liu, H.~Zhou, J.~Wang, and M.~Long, ``Timesnet: Temporal 2d-variation modeling for general time series analysis,'' in \emph{The Eleventh International Conference on Learning Representations, {ICLR} 2023, Kigali, Rwanda, May 1-5, 2023}, 2023.

\bibitem{dai2024periodicity}
T.~Dai, B.~Wu, P.~Liu, N.~Li, J.~Bao, Y.~Jiang, and S.-T. Xia, ``Periodicity decoupling framework for long-term series forecasting,'' in \emph{The Twelfth International Conference on Learning Representations}, 2024.

\bibitem{eldele2024tslanet}
E.~Eldele, M.~Ragab, Z.~Chen, M.~Wu, and X.~Li, ``Tslanet: Rethinking transformers for time series representation learning,'' \emph{arXiv preprint arXiv:2404.08472}, 2024.

\bibitem{tonekaboni2021unsupervised}
S.~Tonekaboni, D.~Eytan, and A.~Goldenberg, ``Unsupervised representation learning for time series with temporal neighborhood coding,'' \emph{arXiv preprint arXiv:2106.00750}, 2021.

\bibitem{bai2018empirical}
S.~Bai, J.~Z. Kolter, and V.~Koltun, ``An empirical evaluation of generic convolutional and recurrent networks for sequence modeling,'' \emph{arXiv preprint arXiv:1803.01271}, 2018.

\bibitem{zheng2024parametric}
X.~Zheng, T.~Wang, W.~Cheng, A.~Ma, H.~Chen, M.~Sha, and D.~Luo, ``Parametric augmentation for time series contrastive learning,'' \emph{arXiv preprint arXiv:2402.10434}, 2024.

\end{thebibliography}
% \color{red}
% IEEE conference templates contain guidance text for composing and formatting conference papers. Please ensure that all template text is removed from your conference paper prior to submission to the conference. Failure to remove the template text from your paper may result in your paper not being published. 
\end{document}